\documentclass[10pt,twocolumn,letterpaper]{article}

\usepackage[most]{tcolorbox}
\usepackage{booktabs} % Add this in your preamble
\usepackage{multirow}
\usepackage{tabularx}    % for flexible-width tables
\usepackage{wacv}              % To produce the CAMERA-READY version
\usepackage{makecell}
\usepackage{tabularx}
\usepackage{float} % in preamble

\definecolor{wacvblue}{rgb}{0.21,0.49,0.74}
\usepackage[pagebackref,breaklinks,colorlinks,allcolors=wacvblue]{hyperref}

\def\wacvPaperID{17} % *** Enter the WACV Paper ID here
\def\confName{WACV}
\def\confYear{2026}

\title{From Filters to VLMs: Benchmarking Defogging Methods through Object Detection and Segmentation Performance}

\author{
Ardalan Aryashad\thanks{Equal contribution authors.}\footnotemark[1] ,
Parsa Razmara\footnotemark[1] ,
Amin Mahjoub\footnotemark[1] ,\\
Seyedarmin Azizi\footnotemark[1] ,
Mahdi Salmani\footnotemark[1] ,
Arad Firouzkouhi\footnotemark[1]\\
Los Angeles, CA, 90089 \\
University of Southern California \\
{\tt\small \{aryashad, prazmara, amahjoub, seyedarm, salmanis, firouzko\}@usc.edu}
}

\begin{document}
\allowdisplaybreaks
\maketitle
\begin{abstract}
Autonomous driving perception systems are particularly vulnerable in foggy conditions, where light scattering reduces contrast and obscures fine details critical for safe operation. While numerous defogging methods exist, from handcrafted filters to learned restoration models, improvements in image fidelity do not consistently translate into better downstream detection and segmentation. Moreover, prior evaluations often rely on synthetic data, leaving questions about real-world transferability. We present a structured empirical study that benchmarks a comprehensive set of pipelines, including (i) classical dehazing filters, (ii) modern defogging networks, (iii) chained variants (filter$\rightarrow$model, model$\rightarrow$filter), and (iv) prompt-driven visual--language image editing models (VLM) applied directly to foggy images. To address the gap between simulated and physical environments, we employ both the synthetic Foggy Cityscapes dataset and the real-world Adverse Conditions Dataset with Correspondences (ACDC). 
Crucially, we examine the generalizability of these pipelines by evaluating performance on synthetic fog and real-world conditions. We assess both image quality and downstream performance in terms of object detection (mAP) and segmentation (PQ). Our analysis identifies when defogging is effective, the impact of combining models, and how VLMs compare to traditional approaches. Additionally, to evaluate the quality of dehazed images, we report qualitative rubric-based scores from both VLM and human judges, and discuss their alignment with down-stream task metrics, revealing reasonable correlations with mAP. Together, these results establish a transparent, task-oriented benchmark for defogging methods and highlight the conditions under which pre-processing genuinely enhances autonomous perception in adverse weather conditions. Project Page:  \href{https://aradfir.github.io/filters-to-vlms-defogging-page/}{https://aradfir.github.io/filters-to-vlms-defogging-page/}.
\vspace{-3mm}
\end{abstract}
    
\section{Introduction}
Autonomous driving perception degrades significantly in foggy conditions, where atmospheric scattering reduces contrast and obscures fine details essential for identifying safety critical objects (e.g., pedestrians, vehicles, signage)~\cite{bijelic2020seeing,sakaridis2018foggy,sakaridis2025acdc}. While a wide array of defogging methods exists --ranging from classical dehazing filters~\cite{he2011single,rahman1996multiscale} to modern learning-based models ~\cite{cai2016dehazenet,li2017aod,qin2020ffa}--improvements in visual fidelity do not consistently translate into better downstream perception~\cite{sakaridis2018foggy,li2017aod}. Furthermore, prior evaluations have heavily relied on synthetic benchmarks, leaving critical questions unanswered about their transferability to realistic driving scenarios~\cite{li2019reside}.

\begin{figure*} % figure* spans across both columns
    \centering
    % First subfigure
    \begin{subfigure}[b]{0.32\textwidth}
        \centering
        \includegraphics[width=\linewidth]{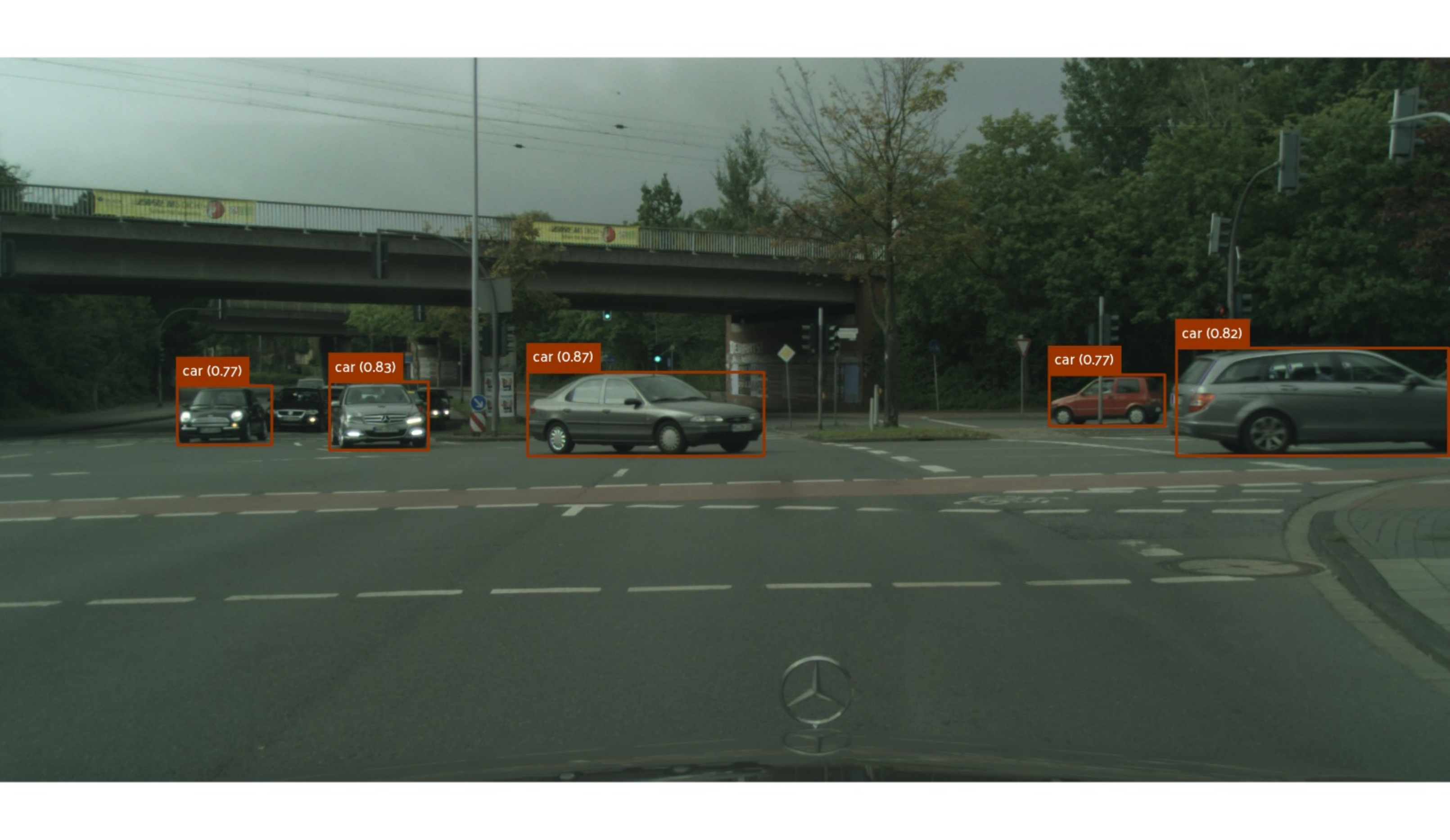}
        \caption{Ground Truth}
        \label{fig:introsub1}
    \end{subfigure}
    \hfill
    % Second subfigure
    \begin{subfigure}[b]{0.32\textwidth}
        \centering
        \includegraphics[width=\linewidth]{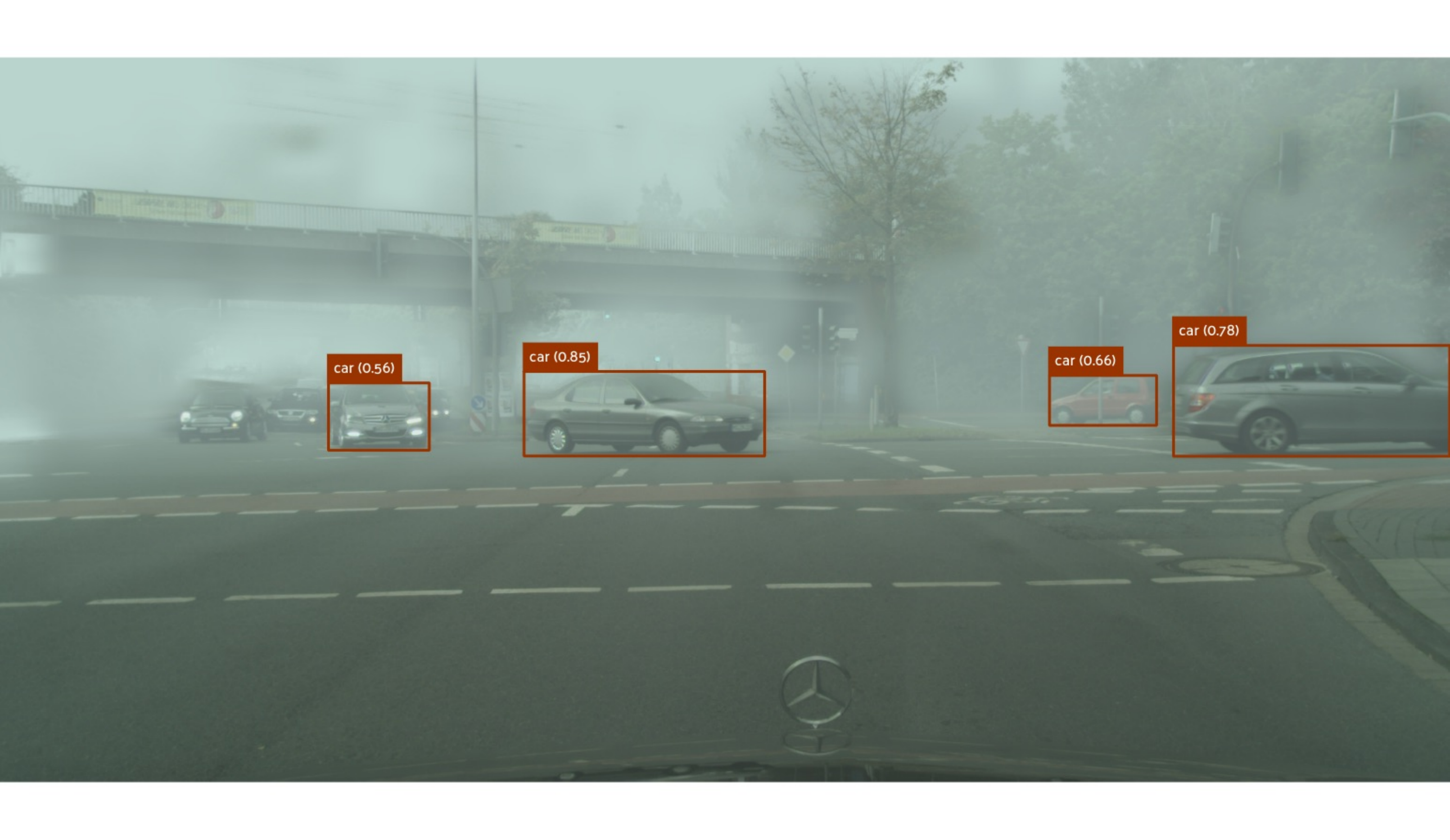}
        \caption{Foggy image}
        \label{fig:introsub2}
    \end{subfigure}
    \hfill
    % Third subfigure
    \begin{subfigure}[b]{0.32\textwidth}
        \centering
        \includegraphics[width=\linewidth]{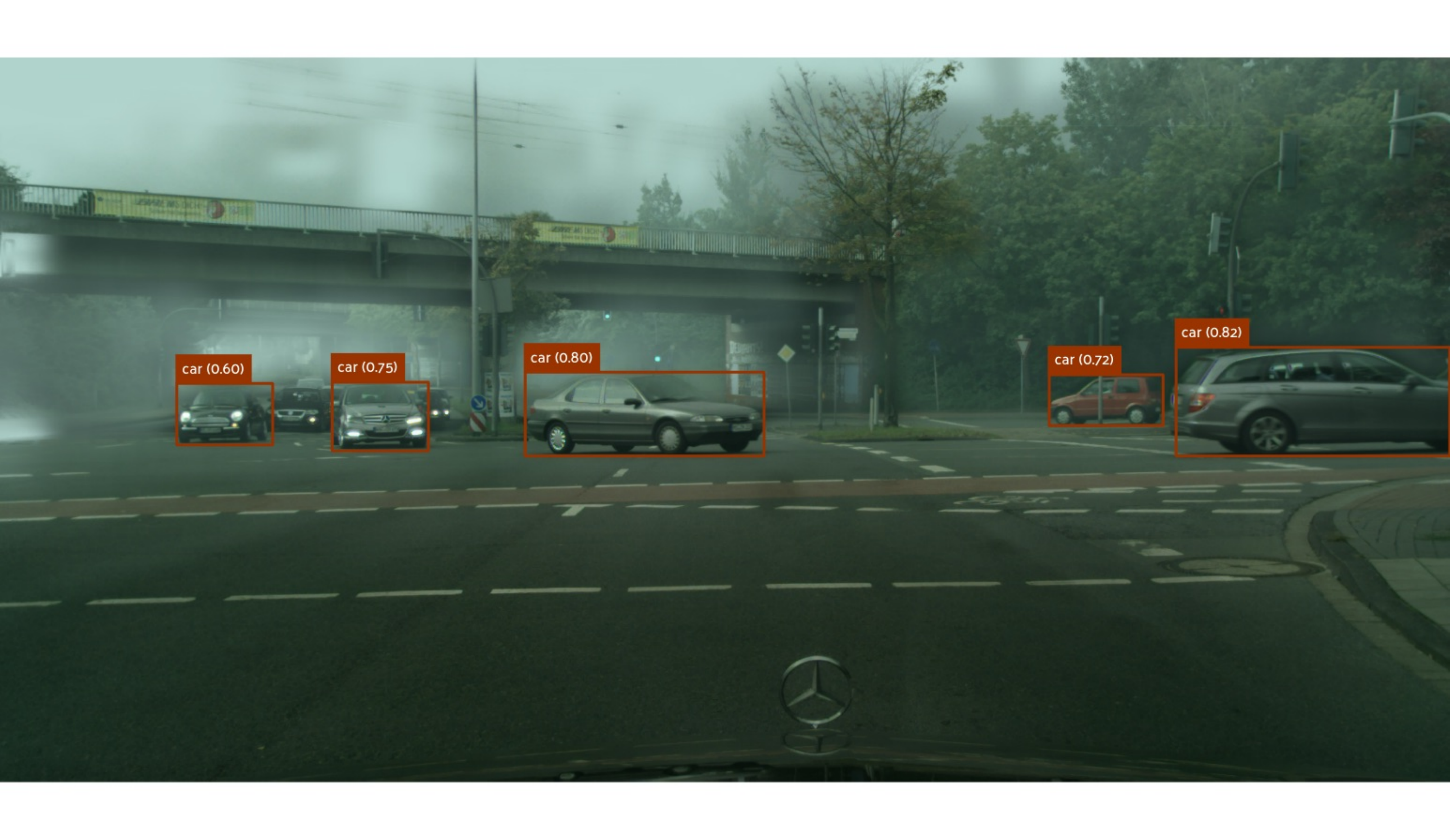}
        \caption{Defogged image}
        \label{fig:introsub3}
    \end{subfigure}
    \caption{Example showing how defogging improves downstream object detection by restoring visibility of fog-obscured objects.}
    \label{fig:intro}
    \vspace{-3mm}
\end{figure*}

To address these gaps, this paper presents a structured empirical study benchmarking a comprehensive set of processing pipelines within a consistent evaluation framework: (i) raw foggy inputs, (ii) handcrafted filters, (iii) learned defogging models, (iv) chained pipelines (\emph{filter}$\rightarrow$\emph{model} and \emph{model}$\rightarrow$\emph{filter}), and (v) prompt-driven image editing model (VLM). Crucially, we extend beyond synthetic evaluation by incorporating the Adverse Conditions Dataset with Correspondences (ACDC)~\cite{sakaridis2025acdc} alongside Foggy Cityscapes~\cite{sakaridis2018foggy}. By training and evaluating on synthetic data and testing on real-world footage, we explicitly examine the transferability of these methods to the real world. We quantify performance not only through image appearance metrics but also via \emph{downstream} object detection and segmentation tasks~\cite{kirillov2019panoptic,cheng2022mask2former}. This approach establishes a standard for evaluating whether image restoration effectively bridges the synthetic to real fog and translates into improved reliability for downstream perception tasks. Figure \ref{fig:intro} illustrates an instance where our pipeline recovers an object previously masked by fog. 

We introduce a Semantic Chain-of-Thought (CoT) defogging prompt for general-purpose image-editing VLMs in image-to-image mode. The prompt emphasizes explicit qualitative cues such as edge acuity, local contrast, color fidelity, artifact suppression, and semantic consistency. Across Foggy Cityscapes, the VLMs with CoT prompt yield consistent mean average precision (mAP) gains over a minimal baseline prompt and improves segmentation quality, while classical filters and simple chains offer mixed benefits. Regarding the qualitative scores, \textbf{edge sharpness} and \textbf{local contrast} show the strongest correlation with the mAP of the downstream detection, indicating that aligning prompts with task-salient visual cues translates into downstream task improvements.

We primarily identify the problem of over-reliance on pixel-level dehazing fidelity scores as proxies for downstream task performance~\cite{li2019reside}; this is a practice that often obscures the true utility of restoration methods in safety-critical scenarios. Our design isolates this factor by holding the detection and segmentation modules fixed and varying only the pre-processing pipeline, creating a controlled environment to evaluate classical filters, learned defogging networks, and mixed chains. As a novel extension to this benchmarking framework, we also incorporate an exploratory analysis of new, prompt-driven image editors (VLMs) ~\cite{huggingface_controlnet_examples,lhoest2022hub,wolf2020transformers}, assessing their viability against dedicated, domain-specific models.

In support of this exploratory track, we utilized and engineered defogging prompt for general-purpose image-editing model. Unlike standard prompting, this approach explicitly cues qualitative attributes such as edge clarity, local contrast, and semantic consistency, while suppressing common failure modes like hallucinations or exposure errors. To bridge the gap between human visual assessment and machine performance, we further employ both VLM-based and human judges to assign qualitative rubric scores to the outputs. This analysis allows us to quantify the alignment between rubric-based assessments and task metrics, revealing that specific visual cues—particularly edge sharpness and local contrast—show strong correlations with mean Average Precision (mAP)~\cite{kirillov2019panoptic}, thereby identifying the specific visual properties that drive downstream improvements.

We summarize our \emph{contributions} as follows:
\begin{itemize}
    \item A systematic, controlled benchmark of handcrafted filters, learned defogging networks, and VLM-based editors across both synthetic (Foggy Cityscapes) and real-world (ACDC) datasets, explicitly evaluating the ``sim-to-real'' gap in downstream detection and segmentation.
    \item An analysis of sequential processing pipelines (\emph{filter}$\rightarrow$\emph{model}, \emph{model}$\rightarrow$\emph{filter}) that identifies specific interactions leading to improvement or degradation relative to single-stage methods.
    \item An exploration of VLM-based image editing for adverse weather, using an engineered prompting strategy designed to restore visual features essential for accurate detection.
    \item A quantitative study of \textbf{VLM metric alignment}, correlating qualitative rubric scores with downstream performance (mAP) to determine when visual restoration reliably translates to task accuracy.
    \item A transparent, reproducible evaluation protocol (datasets, models, detector/segmenter configurations, and metrics) to facilitate future comparative research (details in Sections~\ref{sec:method} and \ref{sec:experiments}).
\end{itemize}

Section~\ref{sec:related} reviews prior work on classical filters, learned defogging, datasets, and task-driven approaches. Section~\ref{sec:method} details our pipelines, metrics, and qualitative VLM assessment rubric; Section~\ref{sec:experiments} covers datasets and evaluation frameworks; Sections~\ref{sec:results},~\ref{sec:discussion} report results and ablations, followed by conclusions in Section~\ref{sec:conclusion}.
\section{Related Work}
\label{sec:related}

\textbf{Classical priors and image processing.}
Early single-image dehazing was dominated by priors and enhancement filters. The Dark Channel Prior (DCP)~\cite{he2011single} remains the most influential, but suffers from haloing, color shifts, and failures in sky regions~\cite{ngo2021visibility}. Other handcrafted approaches, such as Multi-Scale Retinex~\cite{rahman1996multiscale,rahman1998resiliency} and CLAHE~\cite{zuiderveld1994clahe}, improve visibility by boosting local contrast or dynamic range but often amplify noise and distort colors~\cite{hsieh2022variational}. While fast and lightweight, these methods lack physical grounding and can harm downstream tasks performance.

\textbf{Learning-based dehazing.}
CNNs shifted the field from physics-based priors to data-driven restoration. DehazeNet~\cite{cai2016dehazenet} learned transmission maps, while AOD-Net~\cite{li2017aod} enabled joint optimization with detectors. Later works such as GridDehazeNet~\cite{liu2019griddehaze} and FFA-Net~\cite{qin2020ffa} improved spatial adaptivity. However, CNN models trained on synthetic data often generalize poorly to real fog and may introduce perceptual artifacts misaligned with task features.

\textbf{Transformer-based models.}
Recent vision transformers capture long-range dependencies and deliver SOTA dehazing quality. DehazeFormer~\cite{song2022dehazeformer} and follow-ups on frequency modeling and UHD restoration~\cite{zheng2021uhd,zhang2024fotformer} improve perceptual fidelity on synthetic benchmarks. Yet, they still struggle in dense or real-world haze and risk hallucinating textures, raising concerns for safety-critical perception.

\textbf{Datasets and benchmarks.}
RESIDE~\cite{li2019reside} established large-scale evaluation protocols, while Foggy Cityscapes~\cite{sakaridis2018foggy} and ACDC~\cite{sakaridis2025acdc} focus on driving scenarios. Seeing Through Fog~\cite{bijelic2020seeing} extends this to multimodal sensing. Despite their impact, synthetic datasets dominate, and models trained on them often fail to generalize to real fog.

\textbf{Impact on downstream tasks.}
The benefit of dehazing for detection/segmentation is mixed. AOD-Net showed gains when trained jointly with Faster R-CNN~\cite{li2017aod}, but Sakaridis \etal found preprocessing alone can degrade segmentation due to artifacts~\cite{sakaridis2018foggy}. Fusion studies~\cite{bijelic2020seeing} suggest that dehazing is insufficient in severe conditions. Selective enhancement may help foggy scenes but harm clear ones, motivating condition-aware application.

\textbf{Task-driven and generative approaches.}
Task-aware models optimize directly for recognition, including united dehazing-detection pipelines~\cite{li2018evdd}, detection-friendly dehazing~\cite{fan2024friendnet}, and domain adaptation for foggy segmentation~\cite{ma2022both,iqbal2022fogadapt}. While effective, they remain dependent on synthetic supervision and domain heuristics. More recently, diffusion and vision–language models offer flexible prompt-driven “defogging,” but lack physical grounding and often hallucinate content, making them more suitable for augmentation than robust perception. In addition to architectural and training choices, decoding strategies influence reliability: unconstrained sampling can lead to neural text degeneration \cite{Holtzman2020The}, while bounded-entropy schemes such as Top-H Decoding offer a controllable balance between coherence and creativity \cite{potraghloo2025top}.

\textbf{Broader cross-domain advances.}
The challenges highlighted in defogging, robustness under degradation, efficiency constraints, and alignment with downstream task utility, reflect broader trends across machine learning and applied sciences where fidelity-driven optimization alone has proven insufficient. Prior work on efficient and sustainable learning emphasizes evaluating models beyond raw accuracy, spanning model compression \cite{han2016deepcompression} and green low-level vision pipelines \cite{movahhedrad2024green}. Similar evaluation gaps arise in domain-specific settings: in finance and signal processing, vision-language and generative models require careful benchmarking to avoid spurious structure under noisy or incomplete observations \cite{khezresmaeilzadeh2025vista,torabi2025llm, khezresmaeilzadeh2025morfi}. Across domains, performance gains are best assessed when models are evaluated on downstream tasks under realistic, heterogeneous conditions rather than averaged or idealized settings, as shown in clinical time-series prediction using self-supervised and task-driven learning \cite{jin2025novel,abdollahi2025advanced,ashrafi2024optimizing}. Similar failures under averaging appear in wireless interference modeling, physical transport systems, and vision, where ignoring structured heterogeneity or distribution shift leads to unreliable outcomes, reinforcing the need for condition-aware modeling and benchmarking in defogging pipelines \cite{tarzjani2025gnn,petersen2025electrokinetic,geirhos2020shortcut}. Across medical imaging, studies spanning quantitative MRI, clinical outcome analysis, and semi-supervised generative learning further demonstrate that performance under degraded acquisition or limited supervision cannot be reliably inferred from reconstruction quality alone, motivating task-aware evaluation and robust benchmarking practices \cite{razmara2025feasibility,razmara2024prostate,momeni2025hyperperfusion,golkarieh2025semi}. Finally, fairness-aware and trustworthy learning frameworks reinforce that reliable deployment depends on understanding when and how model outputs fail under distribution shift \cite{mehrabi2021bias,fayyazi2025fair}. Collectively, these cross-domain insights motivate systematic, task-centric benchmarking, rather than unconditional preprocessing or fidelity-based ranking, which directly underpins our evaluation of defogging pipelines.
. %. Related principles also appear in physical and networked systems, where transport and inference under structured heterogeneity, such as charge-patterned nanochannels or wireless interference, exhibit nontrivial failure modes when averaged or simplified, underscoring the importance of condition-aware modeling rather than unconditional aggregation \cite{petersen2025electrokinetic,tarzjani2025gnn}. 

%\textbf{Broader cross-domain advances}
%The challenges we identify in defogging, such as balancing efficiency, robustness, and task-driven utility, echo broader trends across machine learning. For example, efficient learning methods have long been central, from model compression \cite{han2016deepcompression} to green approaches for low-level vision \cite{movahhedrad2024green}. Similar principles are seen in domain-specific applications, where vision-language and generative models are applied in finance \cite{khezresmaeilzadeh2025vista} and medical imaging \cite{golkarieh2025semi,razmara2024fever}, and predictive modeling supports advances in healthcare \cite{abdollahi2025advanced,ashrafi2024optimizing}. Finally, fairness-aware methods \cite{mehrabi2021bias,fayyazi2025fair} highlight the need for trustworthy deployment. These parallel efforts emphasize that progress across domains depends not only on fidelity or accuracy, but also on delivering reliable, efficient, and fair outcomes, which motivates our benchmarking of defogging pipelines.

\textbf{Summary.}
Classical priors are fast but brittle, CNN/transformer restorers excel on benchmarks but face real-world gaps, and generic pre-processing is not always task beneficial. Task-driven pipelines and larger real-world datasets appear essential for reliable deployment.

\section{Methodology }
\label{sec:method}
\label{sec:qual_pipeline}

Motivated by the need to measure \emph{downstream task performance} on the dehazed images rather than only pixel fidelity, our methodology fixes the detection/segmentation module and varies only the defogging pipelines. We compare classical filters, modern learned restorers, chained variants (\emph{filter}\,$\rightarrow$\,\emph{model} and \emph{model}\,$\rightarrow$\,\emph{filter}), and VLM image editing models by an engineered prompt. We conduct experiments on the Foggy Cityscapes and ACDC datasets, reporting object detection (mAP) and segmentation (PQ) metrics alongside a qualitative rubric assessing visual features essential for downstream tasks.

This section details three components. Section~\ref{sec:quan_pipeline} (\emph{Quantitative Evaluation Pipeline}) specifies datasets, pipelines, and metrics, holding the evaluator fixed while swapping defogging pipelines. Section~\ref{sec:quali_pipeline} (\emph{Qualitative Evaluation Pipeline}) introduces VLM and human judges with a rubric over edge clarity, visibility restoration, and perceived detectability, then analyzes its alignment with quantitative metrics. Finally, Section~\ref{sec:prompt} (\emph{Prompt Engineering and Model Evaluation}) formalizes our prompt design where each image editor runs in image-to-image mode without training, and for every image we generate two edits that differ only in the prompt: a minimal baseline prompt and an engineered prompt paired. The engineered prompt decomposes the operation (remove fog $\rightarrow$ sharpen boundaries $\rightarrow$ boost local contrast $\rightarrow$ photometric check), explicitly names safety-critical categories (cars, pedestrians, traffic signs, buildings), and enforces natural lighting and realistic colors.

\subsection{Quantitative Evaluation Pipeline}
\label{sec:quan_pipeline}

\subsubsection{Defogging pipeline} 

We utilize the validation split of the Foggy Cityscapes dataset and the fog validation split of ACDC. To ensure a fair comparison between synthetic and real-world conditions, we specifically select medium-intensity fog images from Foggy Cityscapes, closely approximating the visual characteristics of the real fog found in ACDC. Our experimental pipeline includes multiple dehazing approaches: first, we apply each handcrafted filter (DCP, CLAHE, MSR) individually and store the results for subsequent processing. Second, we process the foggy images using dedicated defogging models (DehazeFormer, MITNet, FocalNet) to generate restored images. Furthermore, we evaluate mixed pipelines in two distinct configurations:
\begin{itemize}
    \item \textbf{Filter $\rightarrow$ Model:} Applying a classical dehazing filter first, then feeding the result into a defogging model.
    \item \textbf{Model $\rightarrow$ Filter:} Applying the defogging model first, followed by refinement via a classical dehazing filter.
\end{itemize}
\noindent
Finally, we process the original foggy images using vision-language editing models (e.g., Nano Banana~\cite{nanobanana}, Flux~\cite{batifol2025flux}) driven by prompt engineering. This comprehensive set of evaluations enables us to compare not only image appearance metrics but also downstream detection and segmentation performance under consistent input conditions.

\subsubsection{Object Detection Metric}\label{sec:metrics}

To assess the effectiveness of each defogging method for object detection, we utilize the Mean Average Precision (mAP) metric. mAP is a standard benchmark in the literature that measures the accuracy of object localization and classification across all categories.
It is defined as the mean of the Average Precision (AP) scores calculated for each class over a range of confidence thresholds. In this study, we employ \texttt{YOLOv11l} as the fixed object detector across all evaluations. For every processed image, the model generates bounding boxes, class predictions, and confidence scores. 
We then compute mAP by aligning these predictions with the ground-truth annotations in both the Foggy Cityscapes and ACDC fog split. By comparing the performance on raw foggy images against the defogged outputs, we explicitly quantify the baseline degradation caused by fog and isolate the performance gains contributed by each defogging method.

\subsubsection{Segmentation Metric} 

In addition to object detection, we evaluate segmentation performance with a primary focus on panoptic segmentation~\cite{kirillov2019panoptic}. Unlike semantic segmentation~\cite{long2015fcn}, which assigns class labels to pixels without distinguishing between object instances, or instance segmentation~\cite{he2017maskrcnn}, which predicts separate masks for countable objects only, panoptic segmentation unifies these tasks. It assigns a semantic class to every pixel while identifying individual instances of countable objects (e.g., cars, pedestrians) and labeling amorphous regions (e.g., road, sky) collectively as ``stuff.''

To quantify performance, we report \emph{Panoptic Quality} (PQ), decomposed into \emph{Segmentation Quality} (SQ) and \emph{Recognition Quality} (RQ). Let $TP$, $FP$, and $FN$ denote the counts of true positives, false positives, and false negatives, respectively, determined via one-to-one Intersection-over-Union (IoU) matching with a threshold $\tau$. A true positive corresponds to a matched prediction ground-truth pair where $\mathrm{IoU} \ge \tau$, while unmatched predictions and unmatched ground-truth segments constitute false positives and false negatives. The metrics are defined as follows:
\small
\begin{align}
\mathrm{RQ} &= \frac{TP}{TP+\tfrac{1}{2}FP+\tfrac{1}{2}FN} \\[0.5em]
\mathrm{SQ} &= \frac{1}{TP}\sum_{(p,g)\in\mathcal{M}}\mathrm{IoU}(p,g) \\[0.5em]
\mathrm{PQ} &= \mathrm{SQ} \times \mathrm{RQ}
\end{align}
\normalsize

\noindent
where $\mathcal{M}$ is the set of matched segment pairs. SQ assesses the pixel-level alignment of matched masks, RQ evaluates the model's ability to correctly detect and classify objects, and PQ provides a unified score that improves with both precise boundaries and fewer recognition errors.

\subsection{Qualitative Evaluation Pipeline}
\label{sec:quali_pipeline}

Both Foggy Cityscapes ($2048{\times}1024$) and ACDC ($1920{\times}1080$) feature high-resolution imagery that exceeds the native input dimensions of many generative editing models. Bridging this resolution gap presents a distinct challenge for evaluation. Partitioning images into smaller patch preserves pixel density but often introduces pixel shifts upon recombination, degrading global image coherence. Conversely, downsampling and subsequent up-sampling preserves global structure but introduces interpolation artifacts and coordinate shifts. Since standard object detectors rely on precise pixel-level alignment with fixed ground-truth bounding boxes, these spatial perturbations can artificially penalize mAP scores even when visual clarity is significantly improved. Consequently, to robustly assess restoration performance without the bias of alignment errors, we adopt a qualitative scoring rubric as a critical complement to quantitative metrics. This approach allows us to decouple perceptual improvements —such as enhanced edge clarity and contrast— from the geometric sensitivities of the detection pipeline.

To assess the perceptual quality of defogged outputs, we employ VLM-based and human-based judges. Each assessment instance is structured as a triplet consisting of the candidate defogged image, the corresponding ground-truth clear image, and the original foggy input.
The evaluation rubric is adapted from prior VLM-based assessment studies~\cite{lee2024prometheusvision} and tailored to the specific challenges of adverse weather conditions. The model is prompted to score performance on a 0-5 scale across three distinct axes:

\begin{enumerate}
    \item \textbf{Visibility Restoration:} Evaluates the recovery of visual information across varying depths. A score of 0 indicates no improvement, while a 5 denotes a strong, natural-looking restoration where foreground, mid-range, and background details are clearly legible, effectively simulating the removal of haze.
    \item \textbf{Boundary Clarity:} Assesses the sharpness and definition of object contours and thin structures. This axis penalizes fuzzy or misaligned borders (scores 0-2) and rewards crisp, well-aligned boundaries (scores 4--5) that are critical for precise segmentation.
    \item \textbf{Perceived Detectability:} Estimates the facility with which a generic detector could identify objects in the defogged image. Higher scores indicate that even small or distant objects, previously obscured, are rendered distinct enough for reliable detection.
\end{enumerate}

\noindent
To mitigate the potential biases inherent in VLM judgments, we conducted a parallel human evaluation study. Human annotators evaluated a subset of the data using the exact same rubric and scoring criteria. This comparative analysis corroborates the VLM scores, ensuring that the automated judgments align with human perceptual ground truth.

\subsection{Semantic Consistency via Chain-of-Thought Prompting}
\label{sec:prompt}

Image restoration and editing models optimized for perceptual quality can introduce artifacts and distribution shifts that reduce semantic fidelity for downstream object detection, as observed by Sun et al.~\cite{sun2022rethinking}. This phenomenon is a consequence of the fundamental \textit{perception-distortion tradeoff}~\cite{blau2018perception}, where optimizing for realistic textures can compromise signal fidelity. To mitigate this, we move beyond standard imperative prompting and adapt \textit{Zero-Shot Chain-of-Thought (CoT)} reasoning~\cite{wei2022chain, kojima2022large} for the visual domain. While standard prompts treat defogging as a global style transfer task, our approach utilizes \textit{conditional decomposition}: we structure the prompt to explicitly break the restoration problem into a hierarchy of semantic constraints.

This strategy leverages the reasoning capabilities of the underlying text encoder to condition the diffusion process on three critical axes of fidelity: \textit{Scene Consistency} (preserving global geometric layout), \textit{Object Consistency} (maintaining the cardinality and categories of dynamic agents), and \textit{Relation Consistency} (preserving spatial interactions). By decomposing the task into these sequential reasoning steps, we effectively constrain the generative solution space, forcing the model to prioritize the preservation of ground-truth semantics over unconstrained hallucination. We evaluate this against a baseline imperative prompt (see Table~\ref{tab:prompts}) to isolate the impact of this structured reasoning on downstream detection performance (mAP) using our fixed evaluation stack (Sec.~\ref{sec:experiments}).

%\subsection{Task-Oriented Prompt Engineering}
%\label{sec:prompt}

% We investigate whether targeted prompt engineering can steer general-purpose vision-language editing models toward generating defogged outputs that explicitly benefit downstream tasks. Our design strategy aligns prompt content with the key visual attributes identified in our evaluation rubric —specifically \textit{Visibility Restoration} and \textit{Boundary Clarity}— and measures the impact on detection performance (mAP) using our fixed evaluation stack (Sec.~\ref{sec:experiments}).

% We execute the image editing models without fine-tuning. For each input frame, we generate two edited variants that differ \emph{only} in the prompt structure: a minimal baseline prompt and an engineered task-specific prompt. we constructed the task-specific prompt to explicitly target the visual features most critical for detection. The prompt instructs the model to prioritize sharp object boundaries and high local contrast while ensuring the semantic integrity of safety-critical categories (e.g., cars, pedestrians). This approach aims to force the model into a restoration mode rather than a creative mode.
% To ensure reproducibility and standardize comparisons across different editing models, we expose the exact prompt strings in Table \ref{tab:prompts}.

\begin{table}
    \centering
    \begin{minipage}{0.95\linewidth} % Ensures consistent width within the table float
        
        % --- Task-Specific Prompt (Blue Theme) ---
        \begin{tcolorbox}[
          colback=blue!5!white,
          colframe=blue!60!black,
          title=Task-Specific Prompt (CoT),
          fonttitle=\bfseries,
          boxsep=2pt,
          left=4pt,
          right=4pt,
          top=2pt,
          bottom=2pt,
          enhanced,
          sharp corners,
          before upper={\small}
        ]
        \texttt{Remove fog to restore crystal clarity, natural colors, and contrast. Think step by step: Preserve Scene consistency by keeping global geometry and layout intact. Ensure Object consistency so cars, pedestrians, and traffic lights match the reference. Maintain Relation consistency regarding relative positions and interactions. Output the final defogged image.}
        \end{tcolorbox}

        % --- Baseline Prompt (Gray Theme) ---
        \begin{tcolorbox}[
          colback=gray!5!white,
          colframe=gray!50!black,
          title=Baseline Prompt (Minimal),
          fonttitle=\bfseries,
          boxsep=2pt,
          left=4pt,
          right=4pt,
          top=2pt,
          bottom=2pt,
          enhanced,
          sharp corners,
          before upper={\small}
        ]
        \texttt{Remove fog}
        \end{tcolorbox}
    \end{minipage}
    \caption{Prompt Specifications}
    \label{tab:prompts}
    \vspace{-4mm}
\end{table}
\section{Experiments}
\label{sec:experiments}

\subsection{Dataset Preparation}
We conduct experiments on two datasets: Foggy Cityscapes and ACDC. For Foggy Cityscapes, we specifically select the medium fog intensity ($\beta = 0.01$) to ensure comparability with the real-world conditions found in ACDC. While ACDC provides ground-truth annotations for downstream tasks, Foggy Cityscapes requires preprocessing to enable standard object detection benchmarking. We convert its segmentation labels into the COCO instances format by extracting instance masks for relevant dynamic classes (\emph{person, car, truck, bus, train, motorcycle, bicycle}) and computing their corresponding bounding boxes.

\subsection{Defogging Pipelines}

We construct a diverse array of defogging pipelines by combining classical dehazing filters, defogging networks, and generative VLM editors. These methods are evaluated individually and in sequential combinations to test for complementary effects. The specific components utilized in our benchmark include:

\begin{enumerate}
    \item \textbf{Classical Filters:} We employ three standard non-learning methods: \textbf{DCP} (Dark Channel Prior), which estimates atmospheric light based on dark pixel statistics; \textbf{CLAHE} (Contrast Limited Adaptive Histogram Equalization) for enhancing local contrast; and \textbf{MSR} (Multi-Scale Retinex) for dynamic range compression and color restoration.
    
    \item \textbf{Defogging Networks:} We benchmark three dedicated architectures: \textbf{DehazeFormer}, a transformer-based model known for recovering fine textural details; \textbf{FocalNet}, which utilizes focal modulation for context-aware feature aggregation; and \textbf{MITNet}, a lightweight network designed for efficient restoration.
    
    \item \textbf{Generative VLMs:} For prompt-driven defogging, we utilize \textbf{Flux} and \textbf{NanoBanana}. These are large-scale diffusion models applied in image-to-image mode, leveraging the prompt engineering strategies detailed in Section~\ref{sec:prompt} without fine-tuning.
\end{enumerate}

To strictly evaluate synthetic to real fog generalization, we distinguish between the standard \textbf{DehazeFormer} (pretrained on RS-Haze \cite{li2019rshaze}) and \textbf{DehazeFormer Trained}, which we explicitly trained on Foggy Cityscapes images. This distinction allows us to quantify how well performance optimization on synthetic data translates to the real-world adverse conditions in the ACDC dataset.

% ControlNet REMOVED 
% \noindent
% As a training-based baseline VLM, we adopt \texttt{stable-diffusion-xl-base-1.0} as the diffusion backbone for defogging, combined with ControlNet to condition the diffusion process on foggy images. For training, we follow the setup provided in the HuggingFace ControlNet training pipeline \cite{huggingface_controlnet_examples}, using the Foggy Cityscapes training split as the dataset. The corresponding clean images are used as inputs to the diffusion model, and the end-to-end pipeline is trained with the baseline prompt defined in \ref{sec:prompt} as the text input stream. The model is trained for 10 epochs on the entire training split.

\subsection{Detection Evaluation Framework}

To streamline detection evaluation, we utilize the FiftyOne library~\cite{voxel51_fiftyone} to manage datasets and annotations. For every experimental configuration, we execute the \texttt{YOLOv11l} detector to generate bounding boxes, confidence scores, and class predictions. The library subsequently aggregates these outputs against ground-truth labels to compute overall detection metric (mAP) and facilitates rapid qualitative validation through its built-in visualization tools.

\subsection{Segmentation Evaluation Framework}

For segmentation, we employ Mask2Former~\cite{cheng2022mask2former} pretrained on the Cityscapes dataset~\cite{cordts2016cityscapes}. This model integrates a multi-scale pixel decoder with a transformer decoder to predict a fixed set of mask queries, where each query generates a class label and a corresponding spatial mask via masked attention. To ensure reproducibility, we utilize the official implementation and pretrained checkpoints available on the Hugging Face Hub~\cite{wolf2020transformers,lhoest2022hub}. Following inference, we compute performance metrics using the standard COCO evaluation toolkit~\cite{kirillov2019panopticapi}.

\subsection{Qualitative Data Gathering}

To obtain qualitative assessments of our defogged outputs, we employed \texttt{GPT-5} within a VLM-as-a-judge framework. To ensure consistency across evaluations, we utilized a fixed system prompt and strictly enforced the output format for every instance, minimizing variance in the model's reasoning process. Crucially, we applied the identical evaluation rubric across all experimental configurations to guarantee comparable scoring. For each candidate method, we collected qualitative scores on a representative subset of approximately 100 images, averaging the results to derive stable summary metrics for each axis. The detailed analysis of these qualitative findings and their alignment with human judge scores is presented in Section~\ref{sec:results}.
\section{Results}
\label{sec:results}

In Table \ref{tab:results_pipeline}, we present the object detection (mAP) and segmentation (PQ) performance across all defogging pipelines for both the synthetic Foggy Cityscapes and the real-world ACDC datasets. This side-by-side comparison reveals distinct behaviors in how defogging methods generalize from simulated to real-world adverse conditions.

\begin{table}
  \centering
  \setlength{\tabcolsep}{2pt} % Tighten column spacing
  \begin{tabular}{llcccc}
    \toprule
    \multirow{2}{*}{\textbf{Base}} & \multirow{2}{*}{\textbf{Next}} & \multicolumn{2}{c}{\textbf{Cityscapes}} & \multicolumn{2}{c}{\textbf{ACDC}} \\
    \cmidrule(lr){3-4} \cmidrule(lr){5-6}
     & & \textbf{mAP\%} & \textbf{PQ} & \textbf{mAP\%} & \textbf{PQ} \\
    \midrule
    GT & None & \textbf{25.60} & \textbf{100.0} & N/A & N/A \\
    Foggy & None & 22.95 & 57.9 & \textbf{37.32} & \textbf{48.5} \\
    \midrule
    \multirow{4}{*}{\makecell[l]{Dehaze\\Former(Tr)}} 
      & None & \textbf{25.57} & \textbf{74.4} & \textbf{35.63} & \textbf{44.5} \\
      & CLAHE & 25.55 & 70.3 & 35.06 & 38.5 \\
      & MSR & 25.30 & 59.4 & 31.05 & 35.0 \\
      & DCP & 24.52 & 69.9 & 34.17 & 38.0 \\
    \midrule
    \multirow{4}{*}{\makecell[l]{Dehaze\\Former}} 
      & None & 23.79 & 63.0 & \textbf{38.04} & \textbf{48.0} \\
      & CLAHE & \textbf{24.45} & 62.4 & 37.98 & 46.6 \\
      & DCP & 24.09 & \textbf{64.2} & 34.18 & 42.1 \\
      & MSR & 23.36 & 59.4 & 33.89 & 43.9 \\
    \midrule
    \multirow{5}{*}{CLAHE} 
      & None & 23.83 & 58.0 & 37.73 & \textbf{47.0} \\
      & Dehaze(Tr) & \textbf{25.05} & \textbf{69.9} & 34.77 & 40.7 \\
      & Dehaze & 24.10 & 61.1 & \textbf{37.97} & 46.8 \\
      & FocalNet & 23.85 & 60.9 & 37.88 & 46.7 \\
      & MITNet & 23.28 & 56.1 & 32.59 & 46.0 \\
    \midrule
    \multirow{5}{*}{MSR} 
      & None & 23.18 & 50.7 & 33.46 & \textbf{44.5} \\
      & Dehaze(Tr) & \textbf{24.35} & \textbf{53.2} & 32.09 & 31.8 \\
      & Dehaze & 23.43 & 51.5 & \textbf{34.74} & 44.4 \\
      & FocalNet & 23.51 & 51.7 & 33.74 & 44.4 \\
      & MITNet & 23.38 & 57.9 & 33.04 & 34.1 \\
    \midrule
    \multirow{5}{*}{DCP} 
      & None & 23.69 & 64.0 & \textbf{35.08} & \textbf{43.0} \\
      & Dehaze(Tr) & \textbf{23.76} & \textbf{64.8} & 33.14 & 36.1 \\
      & Dehaze & 23.45 & 64.2 & 34.14 & 42.3 \\
      & MITNet & 23.58 & 59.5 & 6.51 & 9.7 \\
      & FocalNet & 23.47 & 63.6 & 34.58 & 42.4 \\
    \midrule
    \multirow{4}{*}{FocalNet} 
      & None & 23.53 & 62.6 & \textbf{37.27} & \textbf{48.0} \\
      & CLAHE & \textbf{23.85} & 60.4 & 37.32 & 45.5 \\
      & DCP & 23.56 & \textbf{62.8} & 34.47 & 41.3 \\
      & MSR & 21.71 & 51.7 & 32.77 & 43.4 \\
    \midrule
    \multirow{4}{*}{MITNet} 
      & None & 22.51 & \textbf{57.9} & \textbf{28.32} & \textbf{31.5} \\
      & CLAHE & \textbf{22.79} & 56.1 & 28.13 & 27.9 \\
      & DCP & 22.38 & 56.8 & 27.89 & 22.7 \\
      & MSR & 21.51 & 51.0 & 22.68 & 25.2 \\
    \midrule
    \multirow{2}{*}{Flux} 
      & CoT Prompt & \textbf{23.76} & \textbf{59.7} & \textbf{37.78} & \textbf{46.6} \\
      & Base Prompt & 23.06 & 55.5 & 29.47 & 43.2 \\
    \midrule
    NanoBanana & Base Prompt & 19.41 & 61.2 & 23.48 & 39.8 \\
    \bottomrule
  \end{tabular}
  \caption{Performance of Defogging Pipelines on Detection (mAP) and Segmentation (PQ). PQ scores for Cityscapes are normalized relative to the ground truth image.}
  \label{tab:results_pipeline}
  \vspace{-6mm}
\end{table}

A critical finding is the \textbf{discrepancy} between trained model on synthetic foggy images and real foggy images. The DehazeFormer (Trained) model, which was explicitly trained on Foggy Cityscapes, achieves near-optimal performance on the Cityscapes validation set (25.57\% mAP), effectively recovering the ground-truth baseline (25.60\% mAP). However, when applied to real fog in ACDC, the same model degrades detection performance (35.63\% mAP) relative to the raw foggy input (37.32\% mAP). This suggests that the model overfits to the uniform, depth-independent noise patterns of synthetic fog, which differ significantly from the heterogeneous scattering found in real environments. 
Conversely, the standard DehazeFormer, pretrained on diverse outdoor datasets, shows better generalization, improving ACDC performance to 38.04\% mAP. \textbf{This illustrates that maximizing metrics on synthetic benchmarks does not guarantee — and may even degrade — performance in real-world deployment.}
Analyzing the chained pipelines reveals that the order of operations significantly impacts downstream accuracy. On Cityscapes, the sequence DehazeFormer $\rightarrow$ CLAHE (24.45\% mAP) outperforms the reverse CLAHE $\rightarrow$ DehazeFormer (24.10\% mAP). We hypothesize that applying the learning-based restoration first removes the bulk of the haze, allowing the subsequent contrast enhancement (CLAHE) to refine edge definitions without amplifying noise artifacts. However, this trend is less consistent in real-world settings; on ACDC, adding post-processing filters often yields diminishing returns compared to the standalone generalist model, showing that real fog is less responsive to basic contrast adjustments.

Classical methods relying on statistical priors show improved performance on synthetic data but fail on real data. For instance, the Dark Channel Prior (DCP) improves Cityscapes detection (23.69\% mAP vs. 22.95\% baseline) but causes a substantial drop in ACDC (35.08\% mAP vs. 37.32\% baseline). This aligns with the understanding that the simplified atmospheric scattering model assumed by DCP holds for synthetic images but breaks down under the complex lighting and non-uniform density of real fog.

Finally, the Flux model driven by our CoT prompt demonstrates that generative editing can competitively restore image features, achieving 23.76\% mAP on Cityscapes and 37.78\% mAP on ACDC. Interestingly, these results fall within a similar range of improvement over the foggy baseline across both datasets. In contrast, the baseline prompt significantly degrades performance on both Cityscapes and ACDC, emphasizing the critical role of prompt engineering. This comparable level improvement across synthetic and real fog likely stems from the fact that these generative models were \textbf{not} explicitly trained on either dataset; thus, they rely on a generalized understanding of defogging that provides a consistent, non-specialized, performance gain regardless of whether the fog is simulated or real.

\begin{figure}
    \centering
    \includegraphics[width=0.95\linewidth]{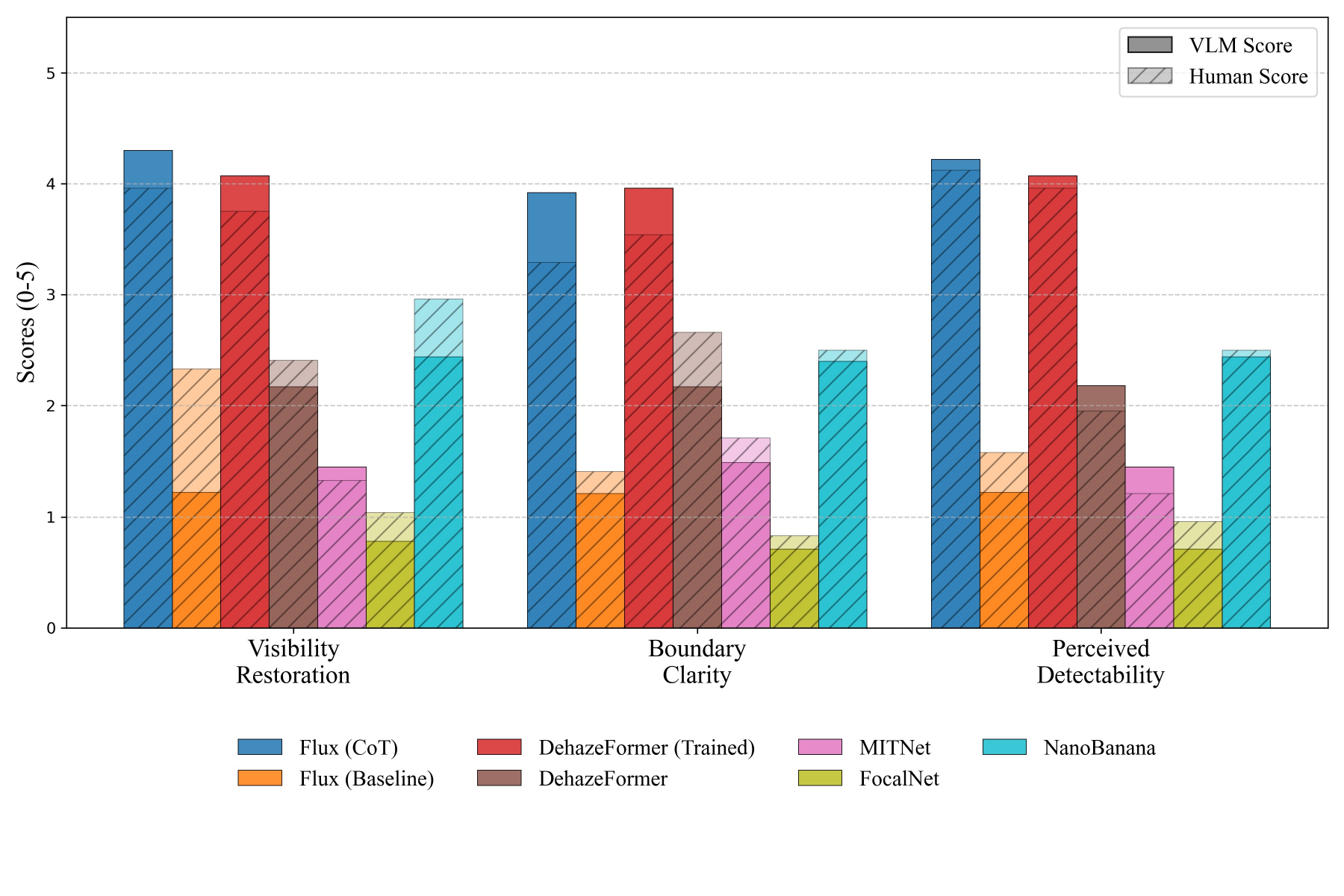}
    \vspace{-4mm}
    \caption{Comparison of the VLM and human judge scores.}
    \label{tab:qualitative-metrics}
    \vspace{-4mm}
\end{figure}

Figure \ref{tab:qualitative-metrics} presents the qualitative assessment of defogging quality as scored by our VLM judge and human judge. The results indicate that the Flux model, when guided by our structured CoT prompt, achieves the highest scores in \textit{Visibility Restoration} (4.30) and \textit{Perceived Detectability} (4.22), outperforming even the specialized DehazeFormer (Trained) model in these perceptual categories. In contrast, the baseline prompt yields significantly lower scores, underscoring the necessity of prompt engineering for effective VLM-based defogging.

\section{Discussions}
\label{sec:discussion}
\begin{figure*}[t]
    \centering
    \setlength{\tabcolsep}{0pt} 
    
    % --- ACDC ROW (REAL WORLD) ---
    \begin{subfigure}{0.49\textwidth}
        \centering
        \includegraphics[width=0.5\linewidth, height=2cm]{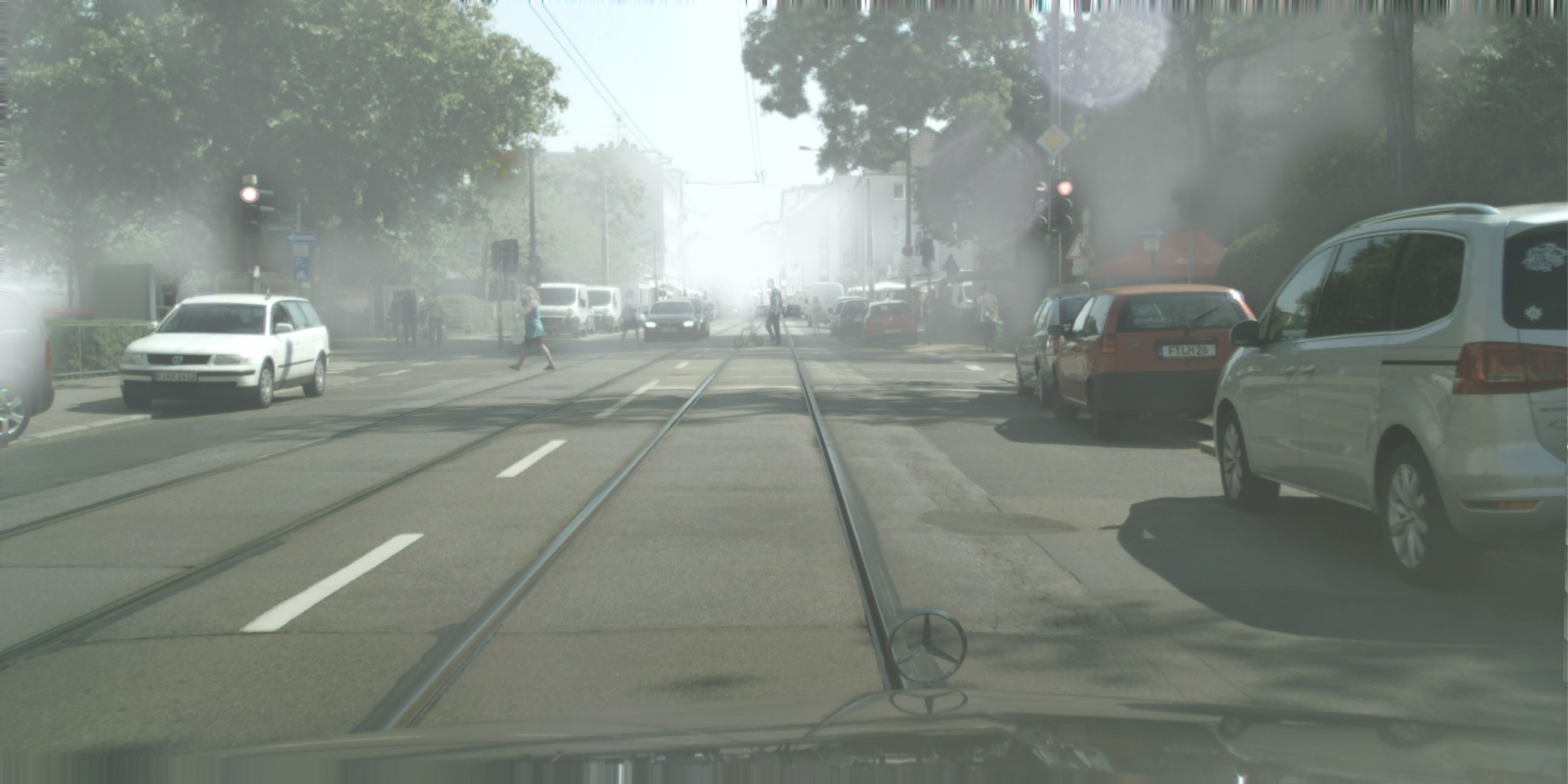}\hspace{0pt}%
        \includegraphics[width=0.5\linewidth, height=2cm]{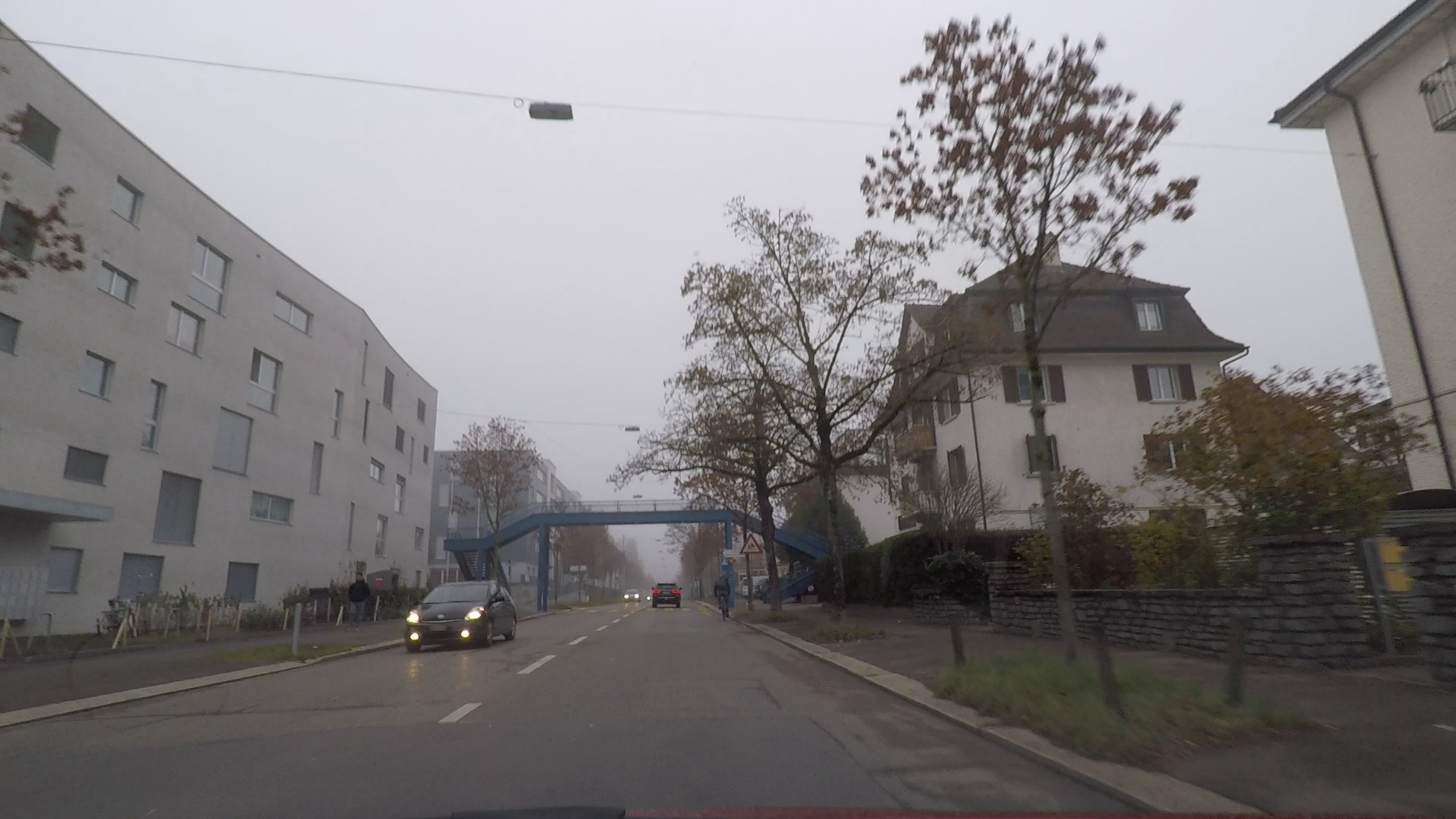} 
        \caption{Foggy Input Images (Foggy Cityscapes - ACDC)}
    \end{subfigure}
    \hfill % Adds the horizontal space in the center column
    \begin{subfigure}{0.49\textwidth}
        \centering
        \includegraphics[width=0.5\linewidth, height=2cm]{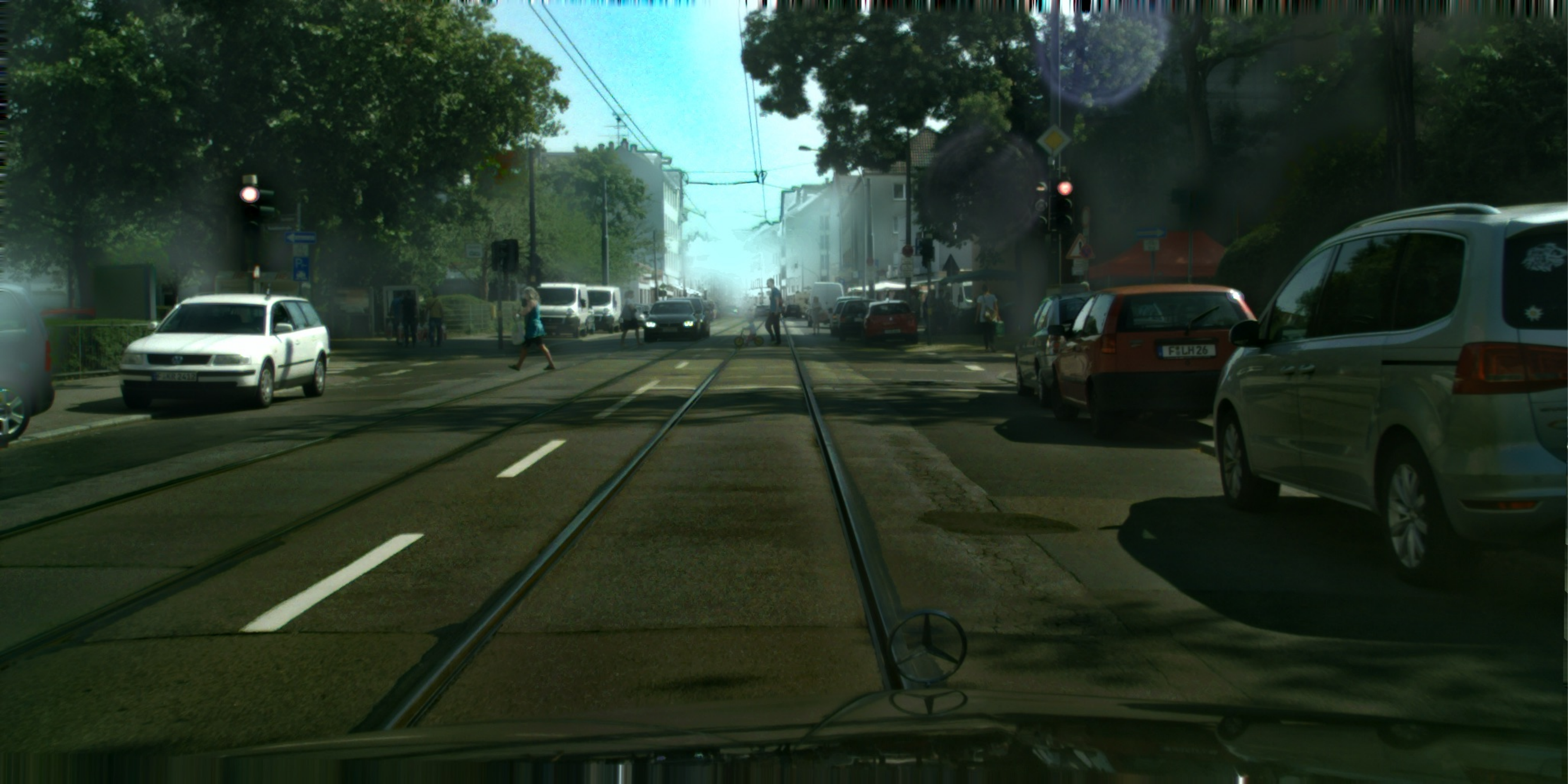}\hspace{0pt}%
        \includegraphics[width=0.5\linewidth, height=2cm]{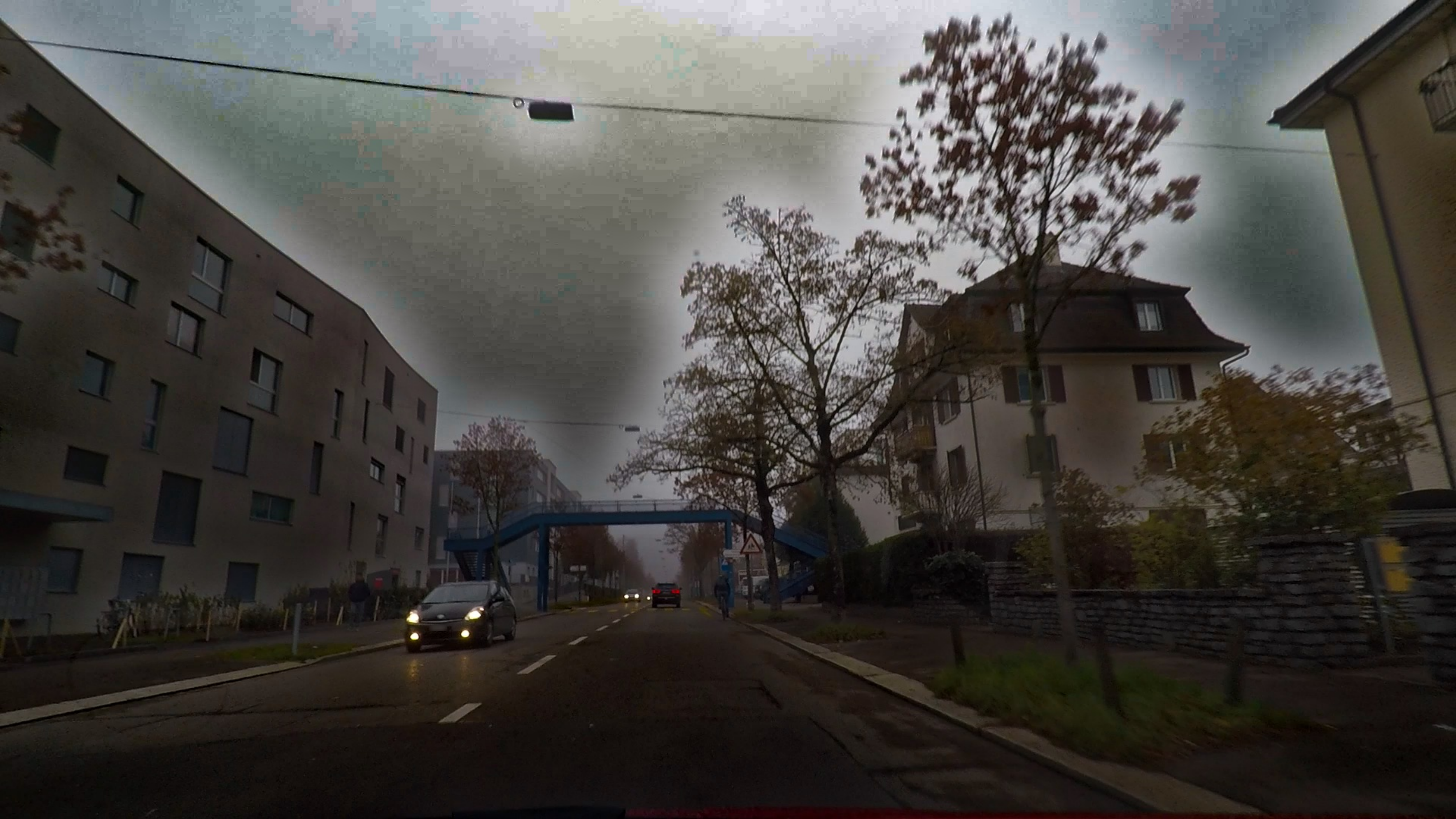}
        \caption{DCP Filter}
    \end{subfigure}

    \begin{subfigure}{0.49\textwidth}
        \centering
        \includegraphics[width=0.5\linewidth, height=2cm]{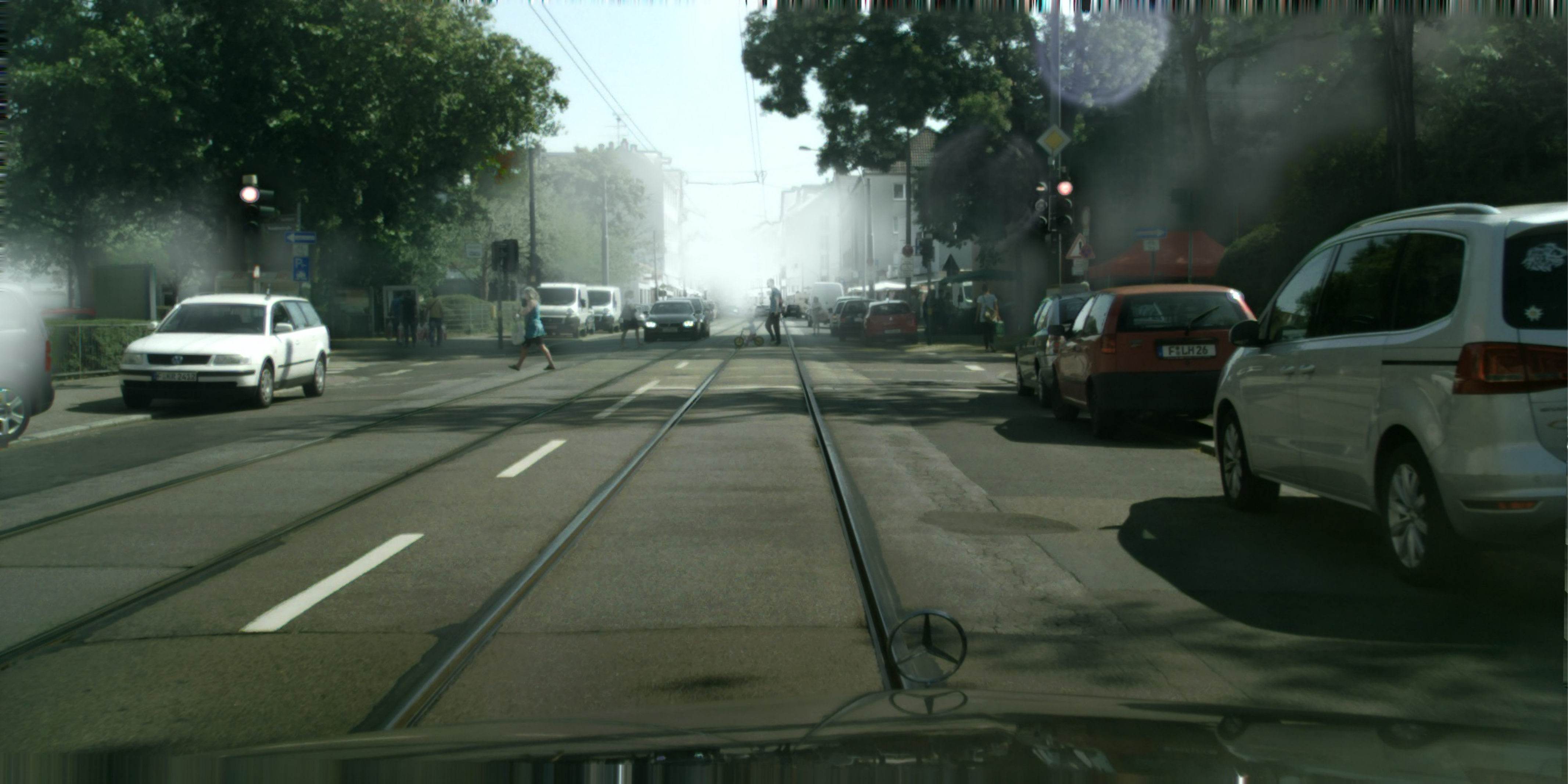}\hspace{0pt}%
        \includegraphics[width=0.5\linewidth, height=2cm]{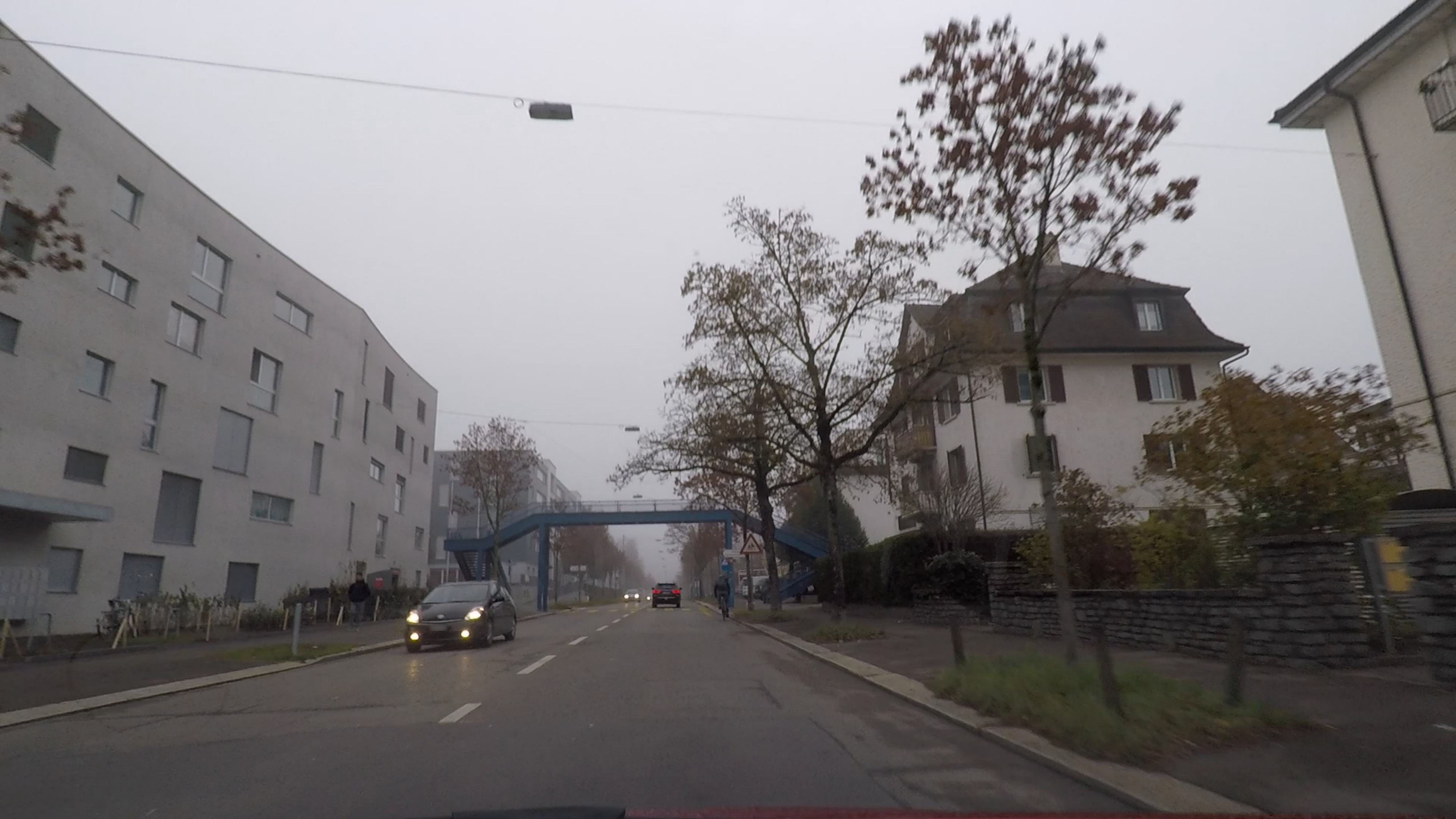}
        \caption{DehazeFormer}
    \end{subfigure}
    \hfill
    \begin{subfigure}{0.49\textwidth}
        \centering
        \includegraphics[width=0.5\linewidth, height=2cm]{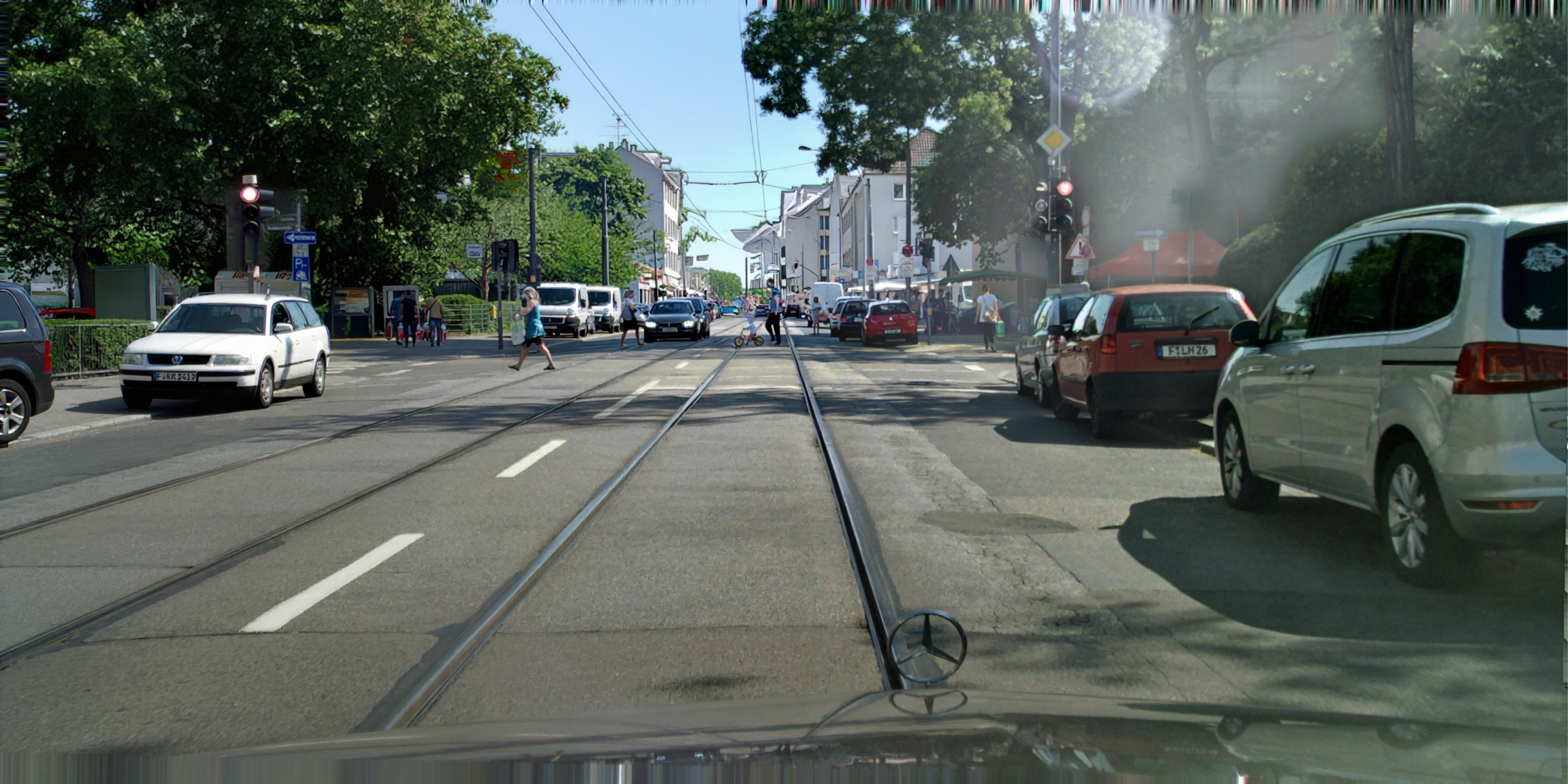}\hspace{0pt}%
        \includegraphics[width=0.5\linewidth, height=2cm]{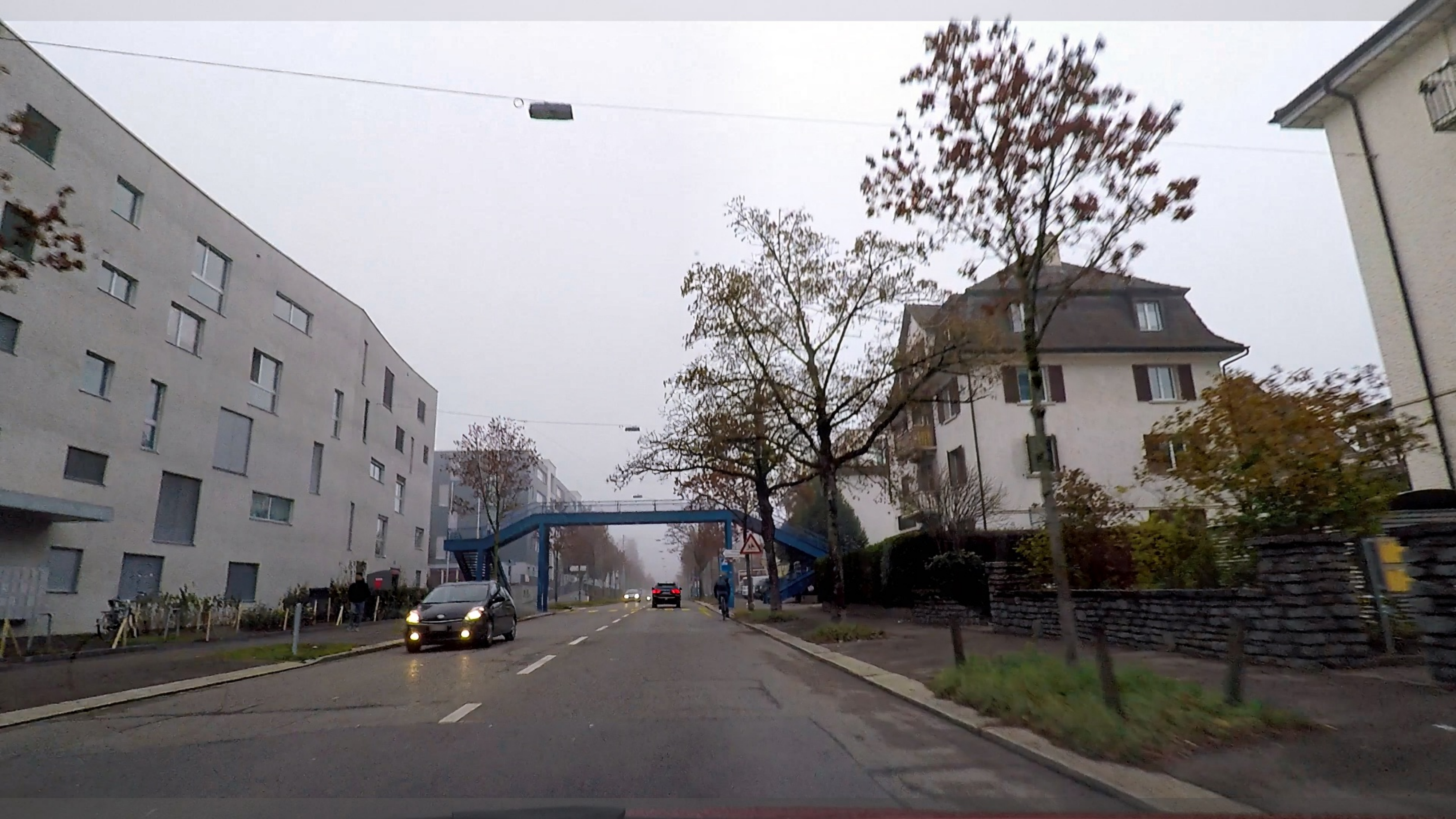}
        \caption{Flux (CoT)}
    \end{subfigure}
    
    % --- CITYSCAPES ROW (SYNTHETIC) ---
    \begin{subfigure}{0.49\textwidth}
        \centering
        \includegraphics[width=0.5\linewidth, height=2cm]{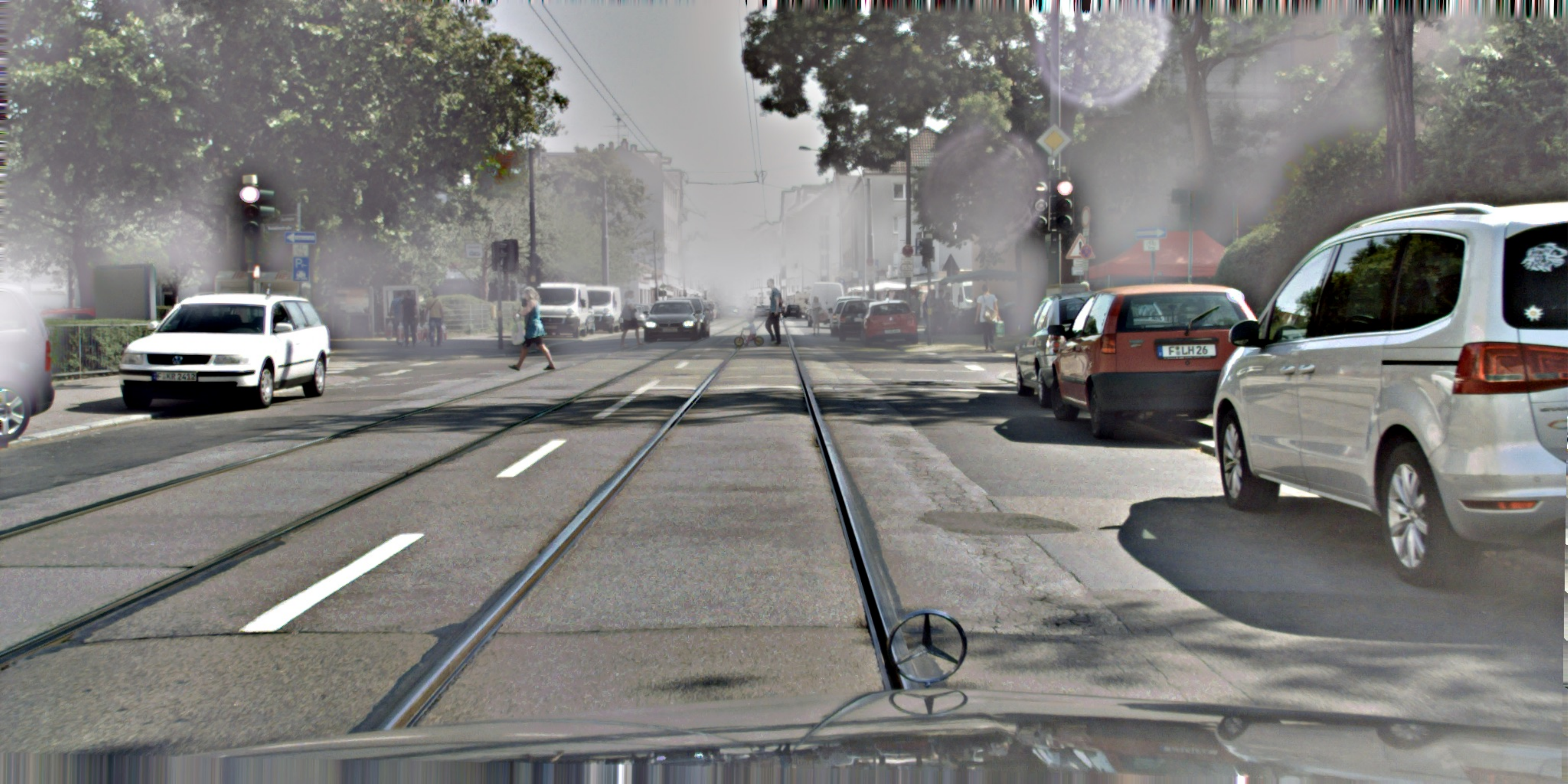}\hspace{0pt}%
        \includegraphics[width=0.5\linewidth, height=2cm]{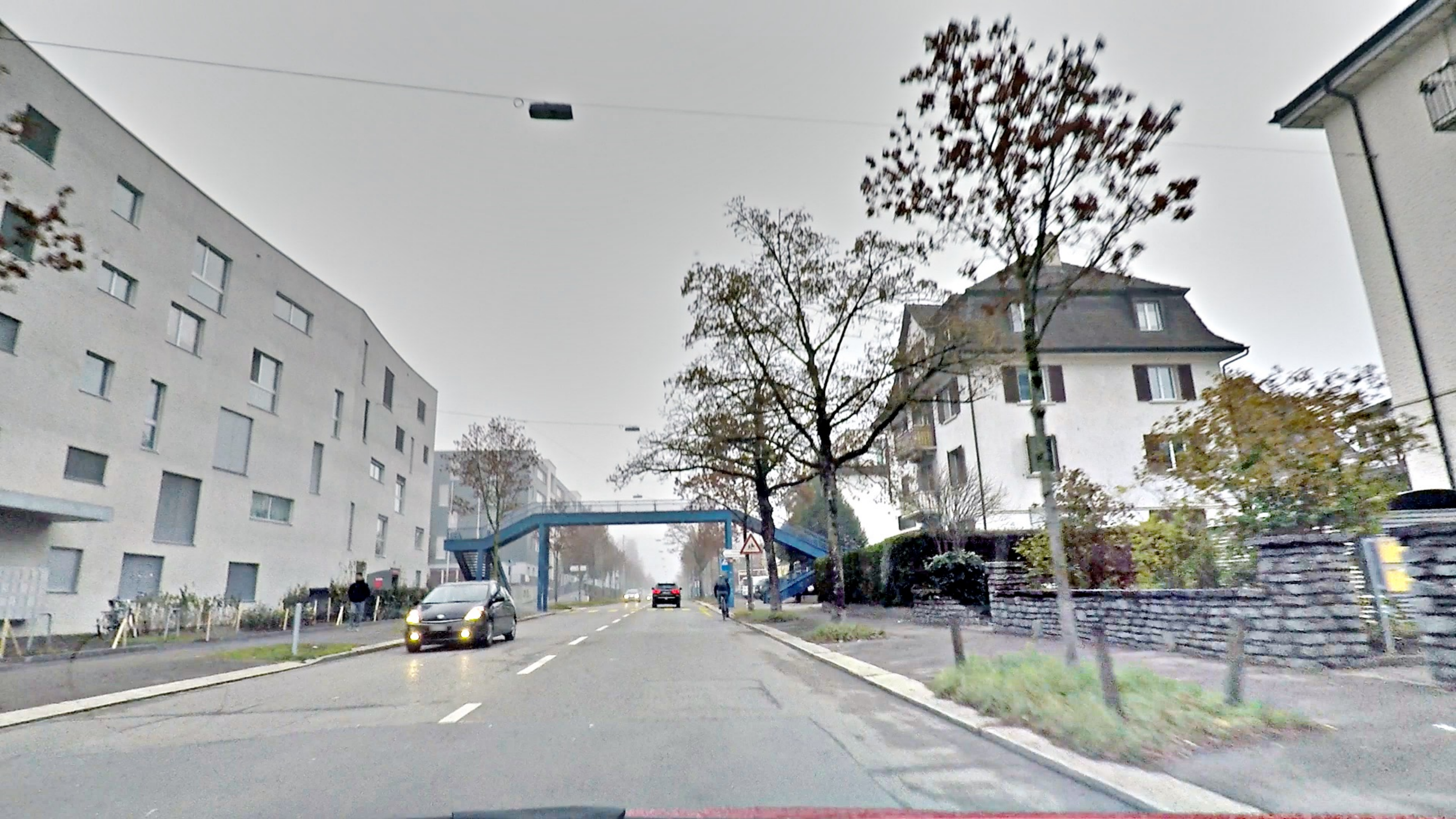}
        \caption{MSR Filter}
    \end{subfigure}
    \hfill
    \begin{subfigure}{0.49\textwidth}
        \centering
        \includegraphics[width=0.5\linewidth, height=2cm]{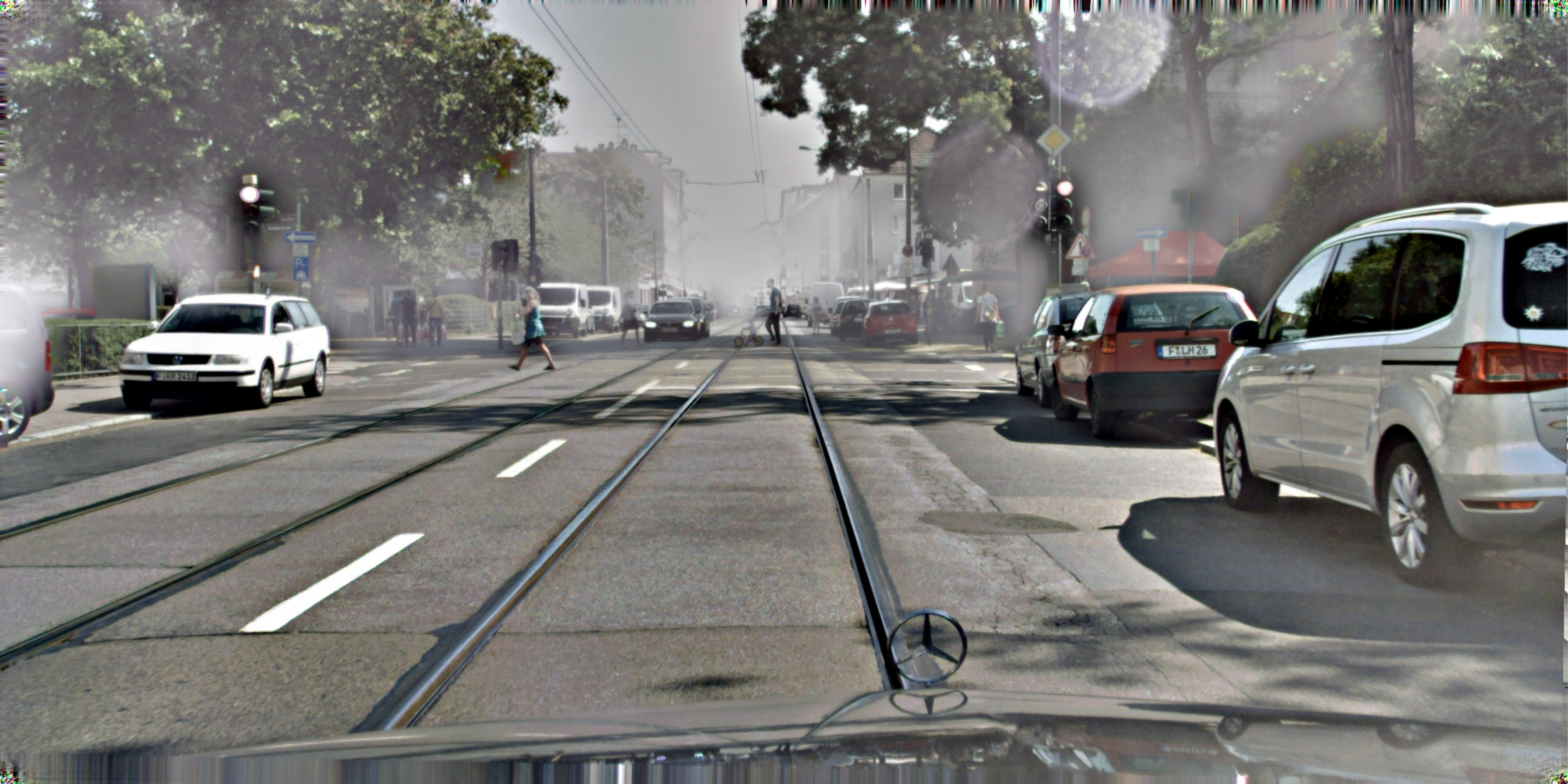}\hspace{0pt}%
        \includegraphics[width=0.5\linewidth, height=2cm]{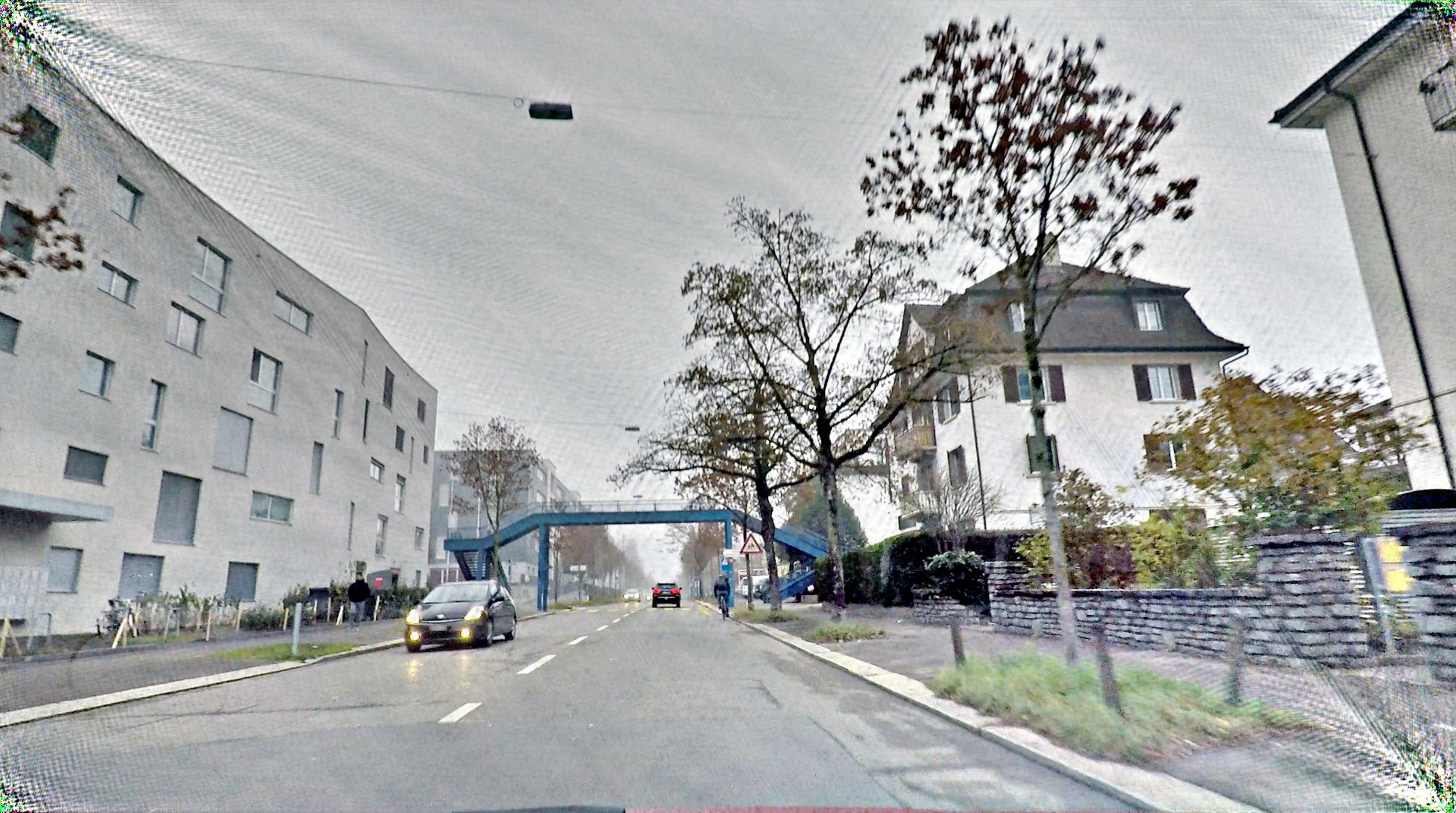}
        \caption{MSR $\rightarrow$ MITNet}
    \end{subfigure}

    \begin{subfigure}{0.49\textwidth}
        \centering
        \includegraphics[width=0.5\linewidth, height=2cm]{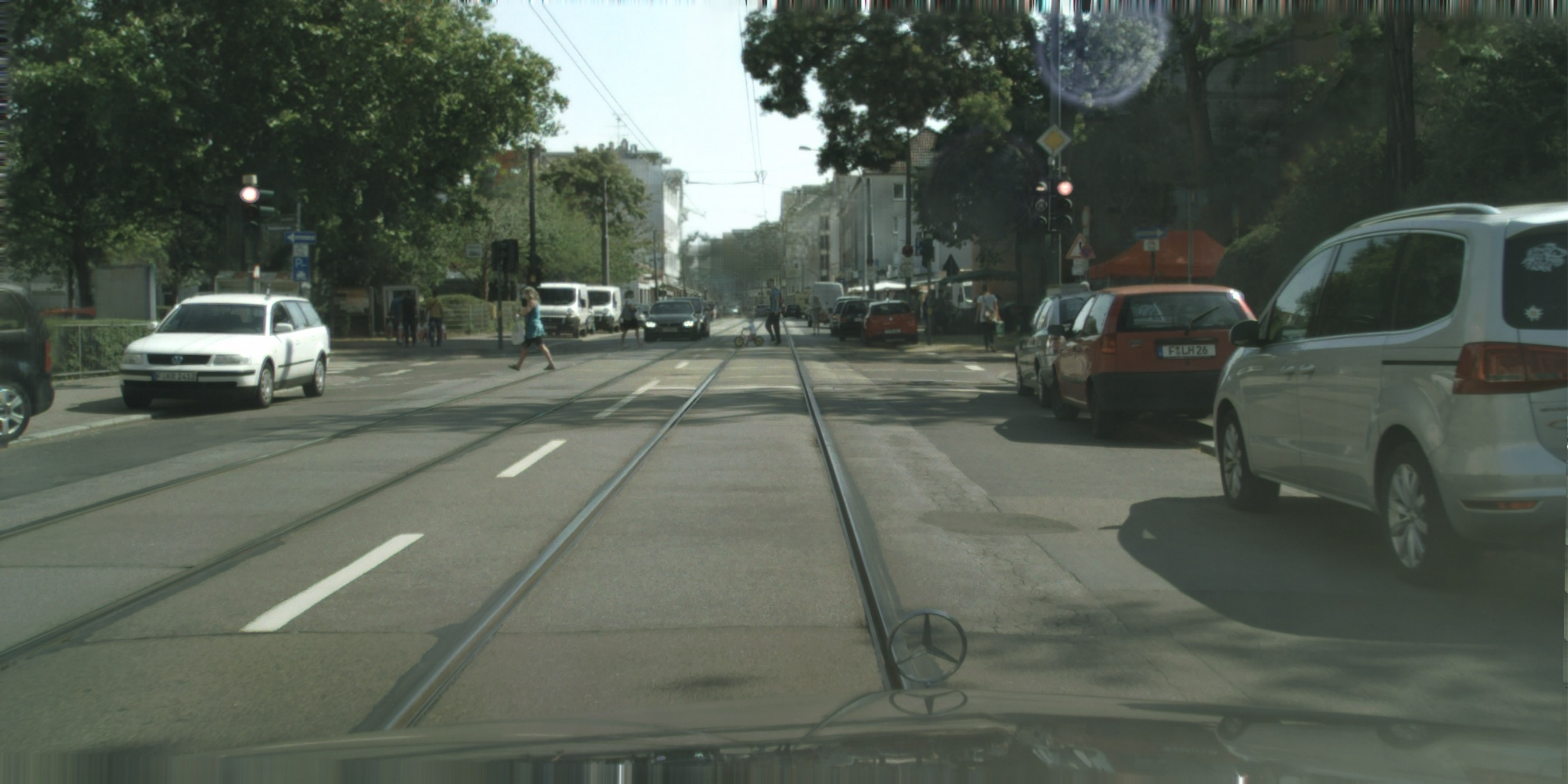}\hspace{0pt}%
        \includegraphics[width=0.5\linewidth, height=2cm]{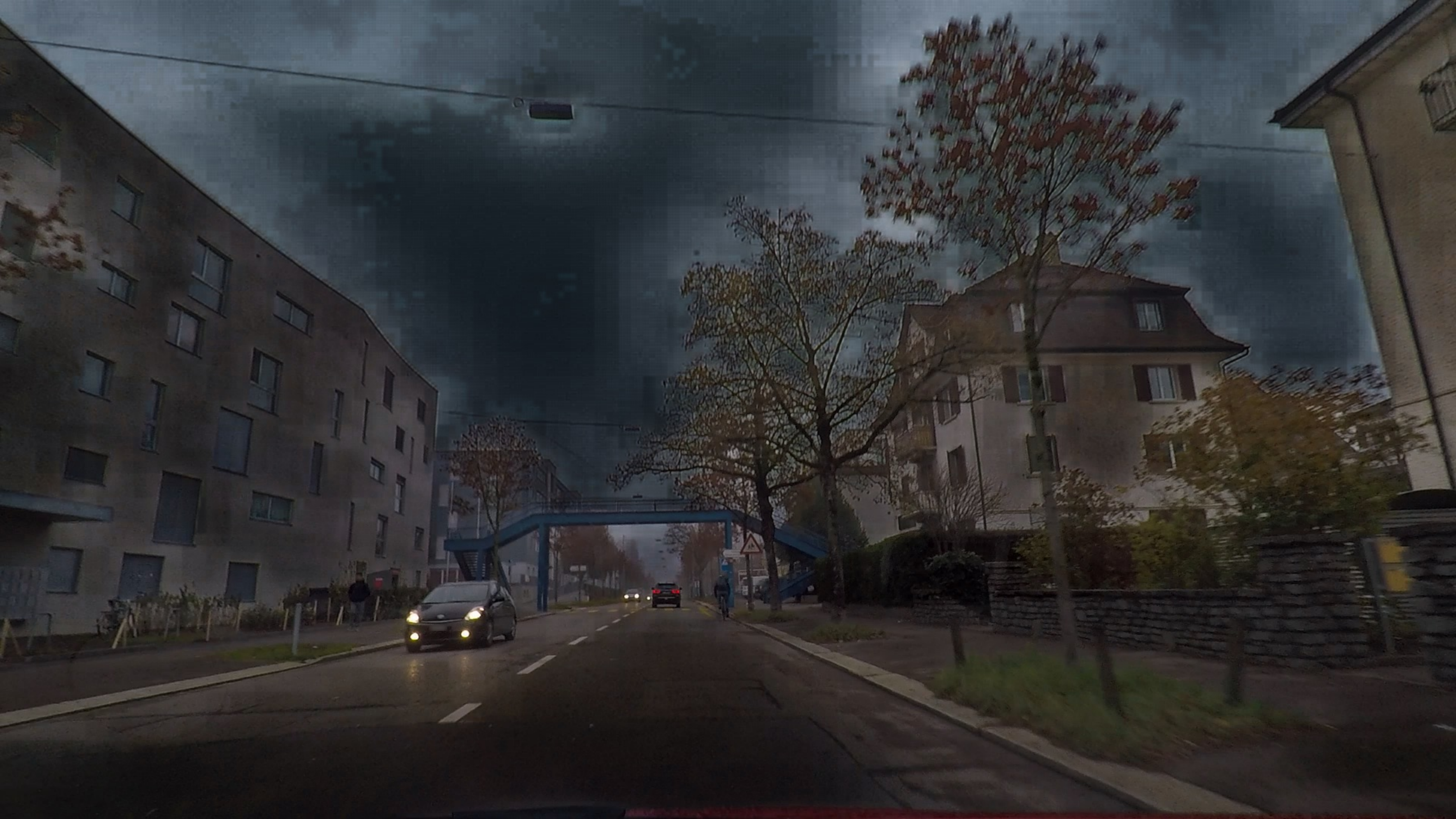}
        \caption{DehazeFormer (Trained)}
    \end{subfigure}
    \hfill
    \begin{subfigure}{0.49\textwidth}
        \centering
        \includegraphics[width=0.5\linewidth, height=2cm]{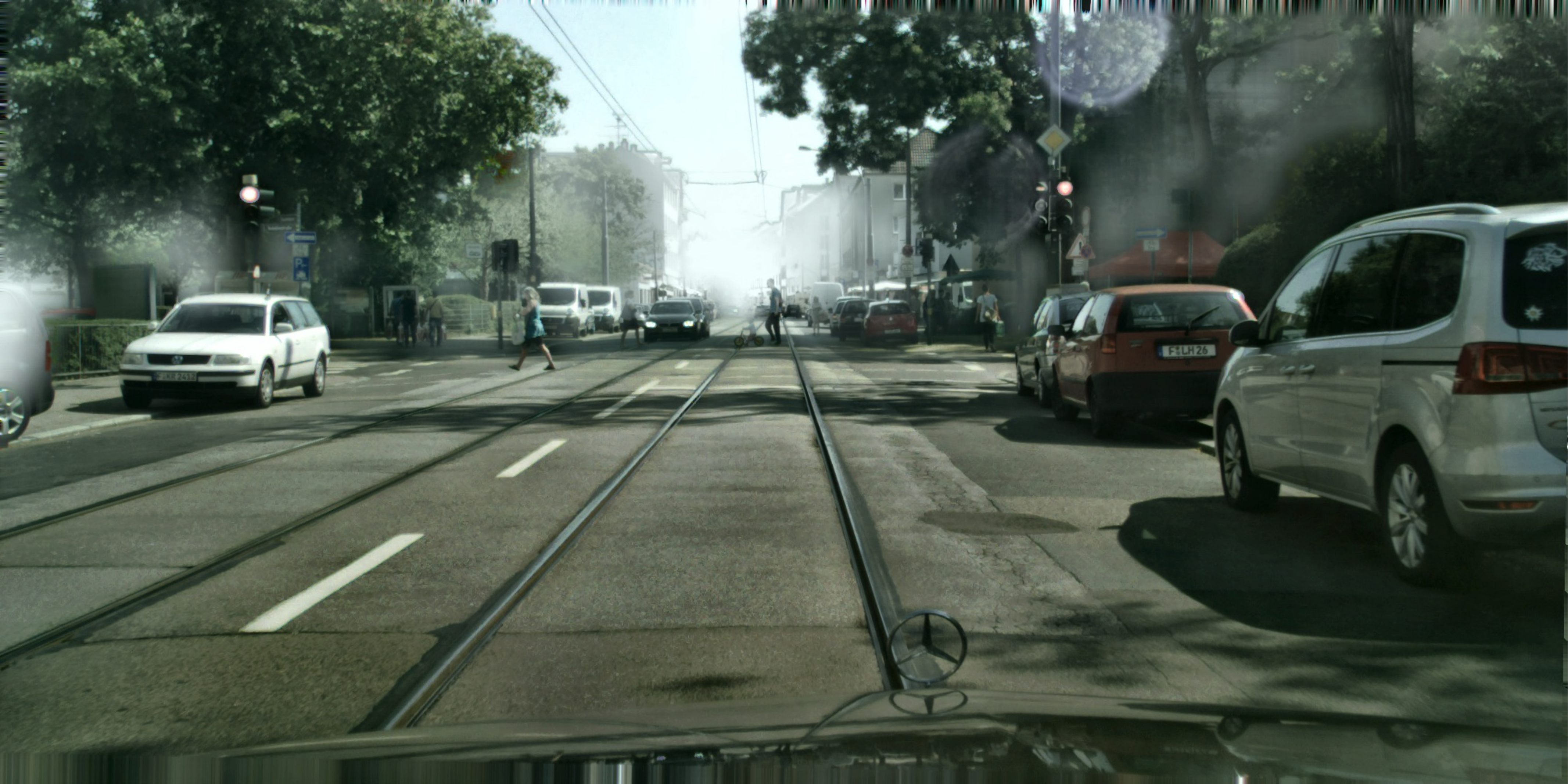}\hspace{0pt}%
        \includegraphics[width=0.5\linewidth, height=2cm]{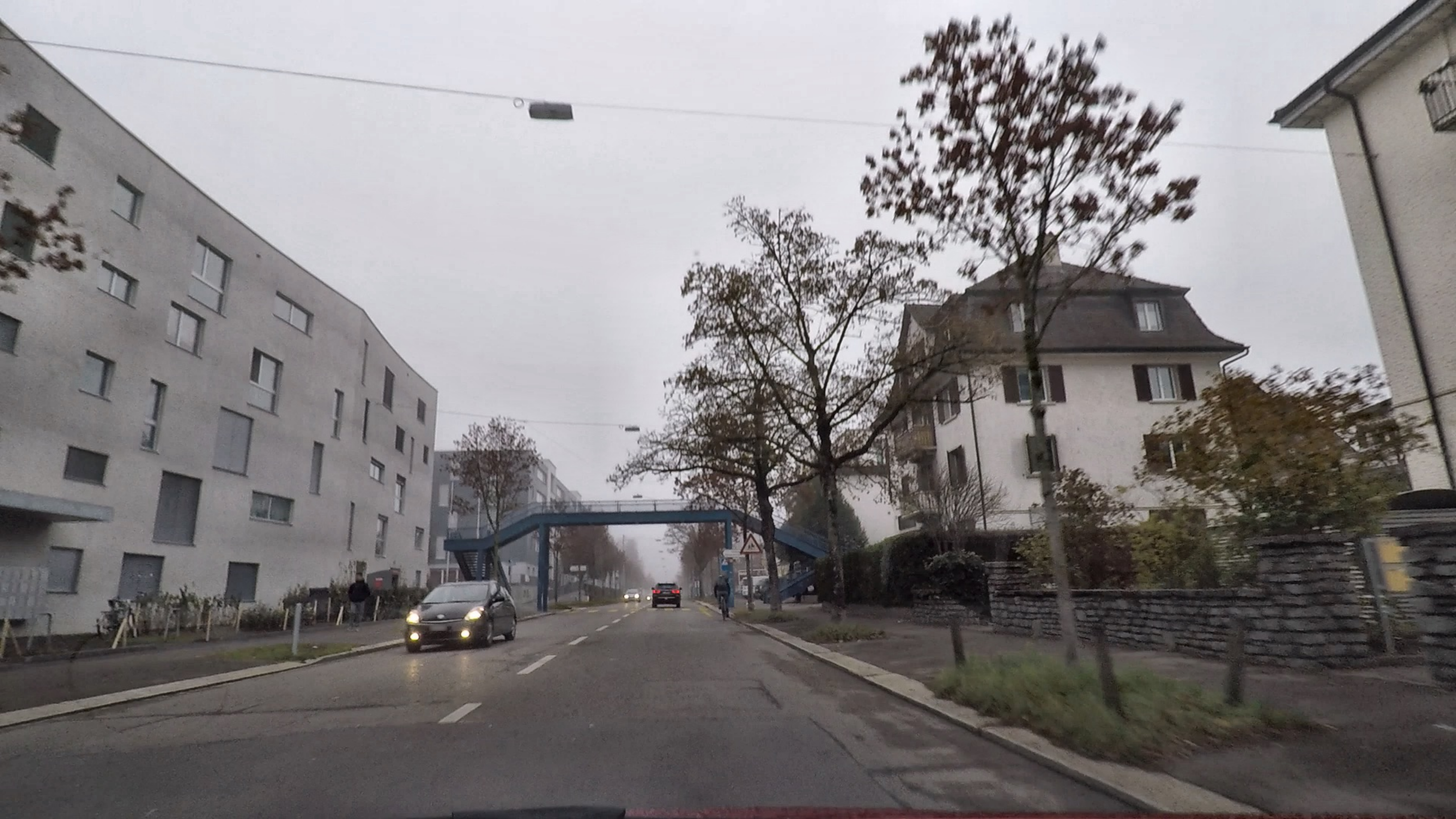}
        \caption{DehazeFormer $\rightarrow$ CLAHE}
    \end{subfigure}
    \vspace{-2mm}
    \caption{\textbf{Comparison across synthetic vs real fog.} (a) shows a sample of input images from Foggy Cityscapes and ACDC dataset. (b-h) 
    Each paired sub-figure demonstrates the output of defogging methods given the input images}
    \label{fig:qualitative_results}
    \vspace{-2mm}
\end{figure*}

\textbf{Qualitative evaluation:} To verify the reliability of our automated VLM-as-judge evaluation, we conducted a parallel human evaluation involving 12 participants who scored the outputs using the same rubric. As shown in Figure ~\ref{tab:qualitative-metrics}, human scores (cross hatch) closely follow the VLM judgments. We computed the Pearson correlation coefficient between the aggregated VLM and human scores, resulting in a strong correlation of $r=0.94$. This high alignment confirms that the VLM judge is not merely hallucinating preferences, but is accurately acting as a proxy for human perceptual ground truth. Interestingly, while the relative ranking of methods remains consistent, the VLM tends to be more penalizing on the baseline prompt (1.22 vs. 1.58 human) and slightly more optimistic about the CoT prompt (4.22 vs. 4.12 human), suggesting a higher sensitivity to the presence or absence of specific visual cues requested in the prompt.

The comparison between Flux (Baseline Prompt) and Flux (CoT Prompt) illustrates the critical role of task-oriented prompt engineering. The baseline prompt ("Remove fog") results in poor \textit{Visibility Restoration} (1.22) and detectability (1.22), performing worse than even the simple NanoBanana model. By contrast, our CoT prompt increases performance to 4.30 for visibility and 4.22 for detectability. This dramatic shift confirms that generic generative models possess the capability for defogging, but this capability remains latent until it is unlocked by a precise, structured prompting that constrains the solution space.

In terms of \textit{Visibility Restoration}, Flux outperforms DehazeFormer (4.30 vs. 4.07), producing images that appear sharper and more effectively cleared from fog. However, in terms of \textit{Boundary Clarity}, DehazeFormer retains a slight advantage (3.96 vs. 3.92).This distinction helps explain the downstream object detection results (Table \ref{tab:results_pipeline}), where DehazeFormer (25.57\% mAP) outperforms Flux (23.76\% mAP). While the VLM excels at the mere dehazing objective, the dedicated model is better at preserving the precise pixel-level edge structures that YOLO detectors rely on for bounding box prediction, suggesting that for machine perception, structural fidelity matters more than noise removal.

\textbf{Generalization Gap:} 
The results in Table~\ref{tab:results_pipeline} highlight a critical gap between the performance in the synthetic foggy dataset and the real-world cases. Specifically, DehazeFormer (Trained), which was trained on the synthetic Foggy Cityscapes dataset, achieves near-perfect recovery of ground-truth performance in that same domain (25.57\% mAP vs. 25.60\% (GT)). However, this high fidelity fails to generalize to the real-world conditions of the ACDC dataset. In ACDC, the same model actually degrades performance (35.63\% mAP) relative to the raw foggy baseline (37.32\% mAP). This drop indicates that the model has overfitted to the specific synthetic fog-generation function rather than learning the complex physical properties of atmospheric scattering.
This sim-to-real gap is further evidenced by the behavior of classical filters such as DCP, MSR, and CLAHE. Although these filters improve the performance of synthetic images, all of them fail to transfer the improvement to the real-world case. These findings suggest that the metric scores on synthetic datasets are often misleading for deployment readiness. Models that effectively remove artificial noise often break down when confronted with real-world fog that does not follow simplified generative assumptions, necessitating benchmarking on real-world datasets like ACDC to ensure robust perception. More examples of the generalization gap are provided in the supplementary materials.

\textbf{Computational Cost: }
Beyond defogging performance, computational efficiency is a critical factor for deployment in safety-critical perception systems. As summarized in Table~\ref{tab:comp_cost}, we observe a significant variance in processing latency across the different categories of defogging approaches.
Classical filters exhibit the lowest latency, with CLAHE achieving a near-instantaneous processing time of 0.055s per frame. However, their utility is limited by poor real-world generalization (Section~\ref{sec:discussion}). Among dedicated defogging networks, MITNet offers the most favorable balance of speed and complexity, maintaining a mean inference time of 0.343s (approximately 2.92 FPS) with only 2.76M parameters. In contrast, the more complex DehazeFormer and FocalNet architectures demand higher GFLOPs, resulting in latencies exceeding 1.2s per frame on an NVIDIA RTX 2080 Ti.
A major drawback of image editing models is their computational bottleneck. Although the Flux and NanoBanana models provide visibility restoration, their iterative nature requires high-performance server resources and several seconds per frame ($>30$s).

\begin{table}
\centering
\begin{tabular}{lcccc}
\toprule
\textbf{Method} & \textbf{Type} & \textbf{Params}  & \textbf{Avg. Time} \\ \midrule
CLAHE & Filter & -- & \textbf{0.055s} \\
DCP & Filter & -- & 0.293s \\
MITNet & Model & 2.76M & 0.343s \\
DehazeFormer & Model & 2.52M & 1.246s \\
FocalNet & Model & 3.74M & 1.250s \\
%Flux (Opt. Prompt) & Editing & 12B & $>30 $s \\ 
\bottomrule
\end{tabular}
\caption{Computational complexity and inference latency for selected defogging methods. Benchmarked on high-resolution $1024{\times}2048$ images using an NVIDIA RTX 2080 Ti.}
\label{tab:comp_cost}
\vspace{-4mm}
\end{table}

\section{Conclusion}
\label{sec:conclusion}
This work presented a systematic evaluation of various defogging pipelines, ranging from classical filters to dedicated defogging networks and generative VLMs. By assessing these methods across both synthetic and real-world datasets, we have highlighted a "sim-to-real" generalization gap. Our results demonstrate that models optimized strictly on synthetic fog fail to translate their performance to physical atmospheric conditions, sometimes even degrading downstream perception metrics compared to raw foggy inputs. Furthermore, We showed that chaining filters with models can yield either complementary benefits or compounding artifacts, depending on the order of operations, and that VLM-based editors provide a flexible alternative but remain inconsistent in downstream performance.

% Our task-oriented benchmark shows that “better-looking” images do not automatically yield better perception. Across Foggy Cityscapes ($\beta=0.01$), a trained DehazeFormer is the most consistently beneficial for downstream detection/segmentation, while classical filters and simple filter$\leftrightarrow$model chains provide mixed value. Prompted VLM editing with a chain-of-thought rubric can be competitive, but its impact is sensitive to edit controls and resolution/aspect constraints that can confound detector metrics. Finally, qualitative scores that emphasize edge acuity and local ROI contrast align strongly with mAP/PQ, offering a practical proxy.

% In the future, this work can be extended to include end-to-end video defogging and measuring temporal stability, and testing models in multi-sensor (RGB + LiDAR or IR) fusions. Another extension is to evaluate models using even further downstream tasks, such as AV crash rate and disengagement/intervention frequency in adverse weather, planner-level safety events (e.g., time-to-collision violations, emergency braking), and closed-loop policy regret measured via high-fidelity simulation and fleet A/B testing, to isolate the effect of small defogging/detection improvements on overall AV safety.
\newpage
{
    \bibliographystyle{ieeenat_fullname}
    \bibliography{main}
}

\end{document}